\documentclass{article}
\usepackage[T1]{fontenc}
\usepackage{iclr2027_conference,times}

\usepackage{amsmath,amsfonts,bm}

\def\eqref#1{equation~\ref{#1}}

\def\1{\bm{1}}

\DeclareMathAlphabet{\mathsfit}{\encodingdefault}{\sfdefault}{m}{sl}
\SetMathAlphabet{\mathsfit}{bold}{\encodingdefault}{\sfdefault}{bx}{n}

\usepackage{hyperref}
\usepackage{url}
\usepackage{graphicx}
\usepackage{booktabs}
\usepackage{capt-of}
\usepackage{wrapfig}
\usepackage{multirow}
\usepackage{float}
\usepackage{placeins}
\usepackage{subcaption}
\title{When Does Depth Help? \\ Understanding Layerwise Inference in Tabular Foundation Models\\[8pt]}

\newcommand{\PaperAuthorBlock}{%
  \begin{tabular}{c}
    Wenwen Gong$^{1}$, Haotian Gu$^{1}$, Zhen Yang$^{2}$,
    Jiaying Ma$^{1}$, Hanqing Zhang$^{1}$ \\
    Zhongshen Wang$^{1}$, Jingjun Wang$^{1}$, Shouyi Xu$^{1}$,
    Wei Wang$^{3}$ \\[4pt]
    {\normalfont\small
      $^{1}$China Agricultural University \quad
      $^{2}$Tsinghua University \quad
      $^{3}$Nankai University}
  \end{tabular}%
}

\newcommand{\PreprintHeader}{Preprint}

\author{\PaperAuthorBlock}

\iclrfinalcopy

\begin{document}

\maketitle
\lhead{\PreprintHeader}

\begin{abstract}

Transformer-based tabular foundation models (TFMs) achieve state-of-the-art performance on small to medium-sized tabular prediction tasks while processing diverse inputs through a shared stack of layers. Recent layerwise analyses identify strong predictive performance in early TFM representations and substantial overlap in computation across layers. However, early predictive accessibility alone does not establish whether later layers are dispensable for every input, and aggregate accuracy can obscure which inputs benefit from additional computation. We study all 24 in-context-learning layers of TabPFN-3 across 272 classification datasets, combining representation geometry, prediction trajectories, truncation, and structural interventions, while also examining related representation dynamics in four additional TFMs. We uncover a pronounced \emph{expansion-compression} trajectory, with stronger late dimensional contraction associated with larger increases in class-prototype separation. A targeted analysis of early-stabilizing samples shows that substantial representational changes persist after predicted classes stabilize, while locally ambiguous queries obtain larger accuracy gains from shallow to late layers than well-separated queries. Repeating a single pretrained block or perturbing layer order reduces predictive performance on average, constraining simple post-hoc substitutions in the frozen model. These findings distinguish predictive accessibility, class stability, and the benefits of continuing computation, supporting an input-dependent assessment of depthwise redundancy in TFMs.

\end{abstract}

\section{Introduction}

Tabular foundation models (TFMs)~\citep{hollmann2023tabpfn,hollmann2025tabpfn,qu2025tabicl,garg2025real,zhang2026mitra,ma2026tabdpt} use in-context learning to predict on new datasets from labeled support examples, without task-specific parameter updates. A single pretrained model processes tasks with different feature spaces and class structures through the same stack of in-context-learning layers. Understanding how much different inputs benefit from this depth is important for assessing computational redundancy and identifying opportunities to reduce inference cost while preserving prediction quality~\citep{elbayad2020depth,fan2020layerdrop}.

Recent layerwise analyses identify strong predictive performance in early TFM representations and substantial overlap in computation across layers~\citep{balef2026one}. TabSwift~\citep{liu2026tabswift} further demonstrates per-query early exiting and observes that deeper exits concentrate in ambiguous regions of representation space. 
These findings motivate examining what subsequent computation still contributes. Early predictive accessibility does not establish that later layers are dispensable for every input, and aggregate accuracy can obscure which inputs benefit from them.

\begin{figure}[!ht]
  \centering
  \includegraphics[width=\textwidth]{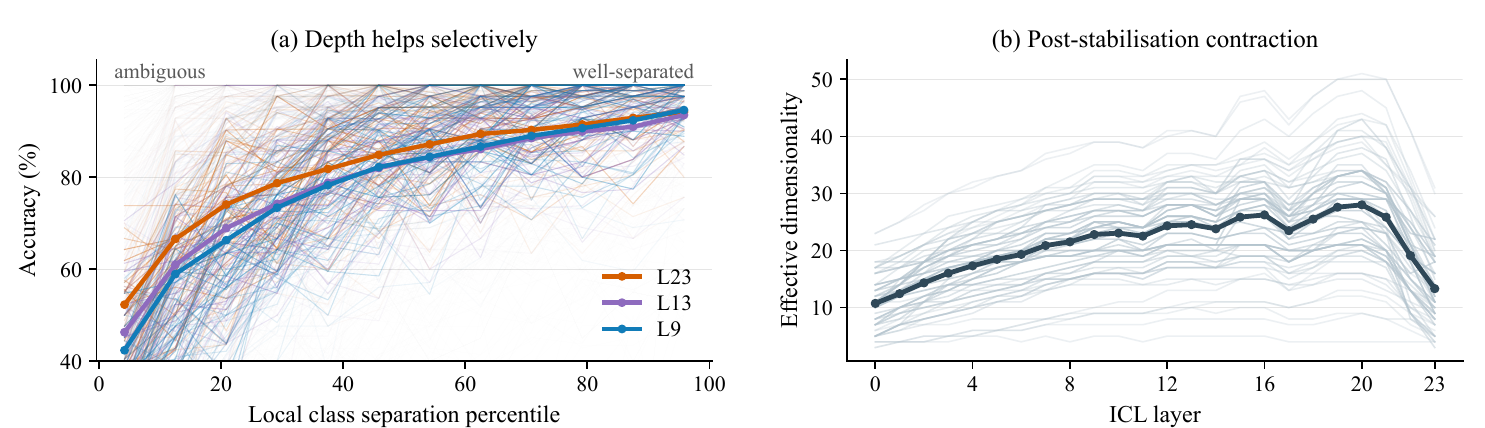}
  \caption{\textbf{Predictive accessibility, class stability, and the value of depth are
  distinct.} (a) L9/L13/L23 accuracy vs.\ local class separation:
  ambiguous queries gain more from depth. (b) Effective dimensionality
  per layer for early-stabilizing queries: representations keep changing after
  the class stabilises, then contract sharply.}
  \label{fig:teaser}
\end{figure}

Our layerwise evaluation of TabPFN-3 makes this distinction concrete. When queries are grouped by local class separation within each dataset, locally ambiguous groups show larger absolute accuracy gains from L9 to L23 on average than well-separated groups (Figure~\ref{fig:teaser} (a)). Meanwhile, a targeted analysis of early-stabilizing samples shows that their representations continue to change substantially after their predicted classes stabilize, including pronounced late dimensional contraction (Figure~\ref{fig:teaser} (b)). These observations highlight the need to distinguish predictive accessibility, class stability, and the accuracy gains from additional computation. They motivate our central question: \emph{When does later computation remain predictively useful, and how does its value vary across inputs?}

We study all 24 ICL layers of TabPFN-3 using 272 classification datasets from OpenML-CC18~\citep{bischl2019openml} and TALENT~\citep{liu2025talent}. We complement its native attention-based retrieval decoder~\citep{grinsztajn2026tabpfn3} with layer-matched classifiers and decoder-independent probes, combining representation geometry, prediction trajectories, truncation, and structural interventions. We also examine related representation dynamics in four additional TFMs~\citep{hollmann2025tabpfn,grinsztajn2025tabpfn25,qu2025tabicl,qu2026tabiclv2}.

We find that TabPFN-3 representations follow a pronounced \emph{expansion-compression} trajectory: early expansion is followed by a relatively stable middle regime and substantial reorganization near the output. Class-relevant information is readable at shallow depth, while class-prototype separation strengthens mainly at later layers. Across datasets, stronger late dimensional contraction is associated with greater increases in prototype separation. These findings distinguish the early availability of predictive information from the subsequent evolution of representation geometry.

At the task level, datasets with later prediction stabilization tend to incur larger accuracy losses under truncation. The overall depthwise accuracy gain also cannot be attributed entirely to mismatch with the final decoder: across 72 OpenML-CC18 tasks, accuracy increases from 82.2\% at L2 to 85.6\% at L23 when each layer is evaluated using its own fitted readout. Additional computation therefore retains predictive value beyond what is accessible through shallow representations under these readouts.

Finally, interventions show that the frozen model is sensitive to changes in its learned layer composition. Replacing distinct blocks with repeated applications of a single pretrained block, or perturbing their execution order, alters representation dynamics and reduces predictive performance on average. These results constrain simple post-hoc substitutions within the pretrained computation. Our findings support evaluating depthwise redundancy through input-dependent predictive gains alongside the accompanying representation dynamics, and motivate future work on adaptive-depth TFM inference.

\section{Related Work}
\label{sec:related}

\noindent\textbf{Tabular Foundation Models.}
Tabular foundation models (TFMs) build on prior-data fitted networks (PFNs)
to perform in-context learning without task-specific parameter updates
\citep{muller2022transformers}. TabPFN-v1 performs ICL over row-level
representations, whereas TabPFN-v2 moves to cell-level representations with
alternating row- and feature-wise attention; TabPFN-2.5 largely retains this
design while increasing depth and scaling to larger tables
\citep{hollmann2023tabpfn,hollmann2025tabpfn,grinsztajn2025tabpfn25}.
In parallel, TabICL introduces a compression-then-ICL architecture that first
constructs fixed-dimensional row representations, with TabICLv2 further
extending this design through repeated feature grouping and target-aware
representations \citep{qu2025tabicl,qu2026tabiclv2}. LimiX expands
structured-data foundation modeling through joint conditional prediction and
retrieval-based ensemble inference \citep{zhang2025limix}, while TabPFN-3
combines feature grouping and row compression in a three-stage architecture
\citep{grinsztajn2026tabpfn3}. Despite this architectural progress,
mechanistic understanding of TFM inference remains limited. Existing studies
have examined spectral adaptivity \citep{zheng2025spectral}, learned
feature-space structure \citep{ye2025closerlook}, intermediate-representation
decodability \citep{hollmann2023tabpfn}, and layerwise predictive
refinement and redundancy \citep{balef2026one}. Our work complements these
studies by jointly examining representation geometry, sample-level prediction
dynamics, and the functional role of depth.

\noindent\textbf{Layerwise Analysis of Transformer Representations.}
Layerwise analysis has been widely used to trace how information and
computation evolve across Transformer depth. Probing and tuned readouts
characterize when task-relevant information becomes accessible
~\citep{alain2017probes,hewitt2019structural,tenney2019bert,rogers2020bertology,belinkov2022probing,belrose2023tuned}, while
representation-similarity and geometric analyses characterize how internal
representations change across layers
\citep{kornblith2019similarity,ansuini2019intrinsic,skean2025layer}.
Complementary intervention-based studies test whether individual layers or
their ordering are functionally consequential
\citep{lad2025,sun2025}. Most directly, \citet{balef2026one} identify early
predictive accessibility, iterative refinement, and substantial depthwise
redundancy in TFMs. We extend this layerwise perspective beyond predictive
readability to characterize how geometry and individual prediction
trajectories evolve, and whether later transformations retain functional
value.

\noindent\textbf{Views of Computation Across Depth.}
Depth has been interpreted as both iterative refinement and structured
representation transformation. Residual networks can implement successive
refinements of existing representations
\citep{jastrzebski2018iterative}, whereas geometric studies report
nontrivial depthwise changes in dimensionality and information structure
\citep{ansuini2019intrinsic,skean2025layer}. Related work on shared
transformations and adaptive-depth inference further asks whether computation
can be repeated, truncated, or selectively allocated across inputs
\citep{dehghani2019universal,lan2020albert,elbayad2020depth,
xin2020deebert,zhou2020patience}. These perspectives leave open whether TFM
depth primarily reflects repetitive refinement or an ordered sequence of
distinct, selectively useful transformations.

\section{Experimental Setup and Analysis Framework}
\subsection{Experimental Setup}

\noindent\textbf{Models.}
We use TabPFN-3 as the primary model for layerwise analysis
\citep{grinsztajn2026tabpfn3}. To assess cross-model generality, we
additionally evaluate TabPFN-v2, TabPFN-2.5, TabICL, and TabICLv2
\citep{hollmann2025tabpfn,grinsztajn2025tabpfn25,
qu2025tabicl,qu2026tabiclv2}. Architectural details are provided in
Appendix~\ref{app:model_architectures}, with cross-model results reported in
Appendix~\ref{app:cross_model_generality}.

\noindent\textbf{Datasets and Evaluation Protocol.}
We evaluate binary and multiclass classification on 72 tasks from
OpenML-CC18 \citep{bischl2019openml} and 200 tasks from TALENT
\citep{liu2025talent}. We restrict the study to classification to maintain
a common output space for analyzing layerwise prediction dynamics and
representation geometry. All models use their standard preprocessing and
inference pipelines. The layerwise representation-extraction protocol is
detailed in Appendix~\ref{app:representation_extraction}.

\subsection{Layerwise Analysis Framework}

\noindent\textbf{Analysis framework.}
We analyze layerwise inference from three complementary perspectives:
representation geometry, sample-dependent prediction dynamics, and the
functional role of depth, building on prior layerwise probing, readout, and
TFM analyses
\citep{alain2017probes,belrose2023tuned,kornblith2019similarity,
balef2026one}.
For query $i$, we denote its hidden representation after layer $\ell$ by
$h_i^{(\ell)}$ and track its representation and prediction across depth.
The representation-extraction protocol is detailed in
Appendix~\ref{app:representation_extraction}.

\noindent\textbf{Predictive readout.} TabPFN-3 differs from earlier TabPFN variants in replacing the fixed-width classification head with an attention-based retrieval decoder \citep{grinsztajn2026tabpfn3}. The decoder predicts directly from the relationship between query and support representations: support representations serve as keys, their one-hot labels as values, and query representations as queries. Because all 24 ICL layers preserve the same 512-dimensional hidden size and the decoder uses layer-independent projections, we can apply the same pretrained model-native decoder at every depth to obtain a consistent predictive readout across layers. To control for possible mismatch between this decoder and intermediate representations, we complement the native readout with layer-matched readouts. Full decoder definitions and validation results are provided in Appendix~\ref{app:native_decoder}.

\noindent\textbf{Representation and functional analyses.}
We characterize representation geometry using effective dimensionality
\citep{ansuini2019intrinsic}, PCA displacement, linear CKA
\citep{kornblith2019similarity}, and cosine similarity, capturing
complementary aspects of dimensionality, movement, global structural
similarity, and sample-wise directional alignment.
We trace sample-level prediction trajectories and apply structural
interventions to test when additional depth becomes functionally consequential
\citep{lad2025,sun2025,balef2026one}. Metric definitions are provided in
Appendix~\ref{app:representation_measurements}, with supplementary
prediction-dynamics and functional analyses in
Appendices~\ref{app:sample_commitment} and
\ref{app:functional_depth}, respectively.

\section{Expansion--Compression Emerges Across Depth}
\label{sec:exp}

\subsection{Representational Expansion Precedes Late Compression}
\label{sec:dimensional_expansion}

We first examine how representation geometry evolves across depth. We measure
this evolution using PCA-based effective dimensionality (ED)
\citep{ansuini2019intrinsic}. For the centered query representation matrix
$\widetilde H^{(\ell)} \in \mathbb{R}^{n \times d}$ with singular values
$\sigma_1 \geq \sigma_2 \geq \cdots$, we define
\[
\scalebox{0.80}{$\displaystyle
\mathrm{ED}(H^{(\ell)}) =
\min \left\{
k :
\frac{\sum_{i=1}^{k}\sigma_i^2}
{\sum_{i}\sigma_i^2}
\geq 0.9
\right\}
$}.
\]
ED measures the number of principal directions required to explain most of
the representation variance. Full details are provided in
Appendix~\ref{app:effective_dimensionality}.

\begin{wrapfigure}{r}{0.35\textwidth}
    \vspace{-0.6\baselineskip}
    \centering
    \includegraphics[width=\linewidth]
    {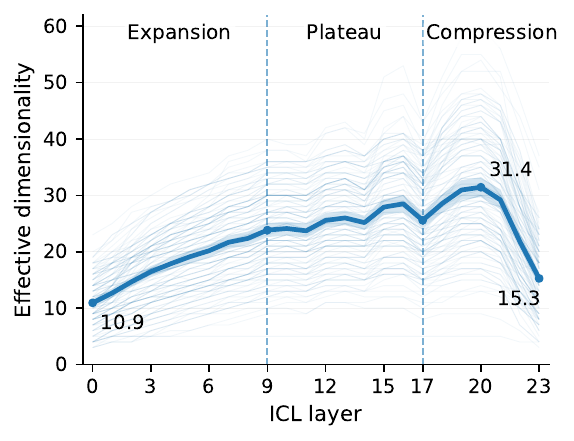}
    \caption{
    \textbf{Representational expansion followed by late compression.}
    Thin curves denote individual datasets and the thick curve denotes
    the dataset-level mean.
    }
    \label{fig:ed_trajectory}
    \vspace{-0.5\baselineskip}
\end{wrapfigure}

Figure~\ref{fig:ed_trajectory} reveals a pronounced non-monotonic trajectory.
Effective dimensionality rises sharply at shallow depth, remains broadly
elevated through the middle layers, and contracts near the output, with a
brief local rebound preceding the final decline. We summarize these stages as
\emph{expansion}, \emph{plateau}, and \emph{compression}; L9 and L17 serve
only as descriptive anchors for subsequent analyses.

This pattern reveals a characteristic geometric trajectory: early layers
broaden the effective representational space, whereas substantial contraction
emerges only near the output. We refer to this trajectory as
\emph{Expansion--Compression}. The pattern persists across retained-variance
thresholds, normalization and covariance controls, and nearby regime
boundaries. Random-weight initialization produces a qualitatively different
depthwise profile
(Appendix~\ref{app:geometry_robustness};
Appendix~\ref{app:random_init_control}).

This late contraction occurs alongside continued representational change.
We therefore next examine how strongly representations are reorganized
within and across these regimes.

\subsection{Late Compression Involves Concentrated Reorganization}
\label{sec:regime_dynamics}

\begin{figure}[!ht]
    \centering
    \includegraphics[width=0.30\textwidth]
    {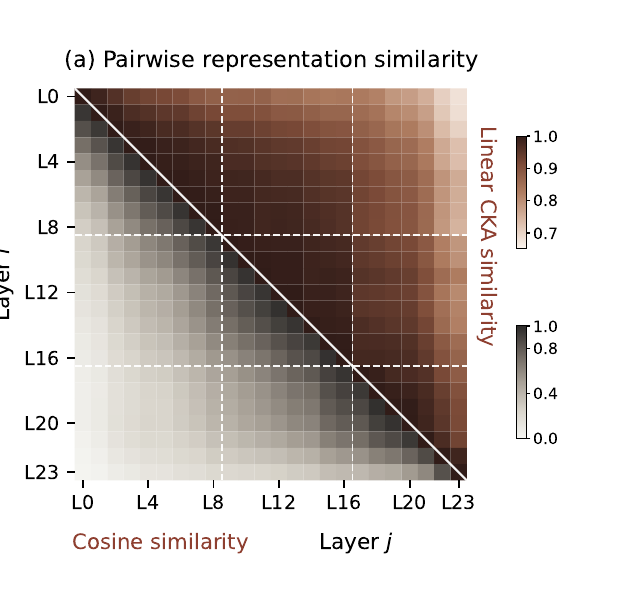}
    \hspace{0.015\textwidth}
    \includegraphics[width=0.31\textwidth]
    {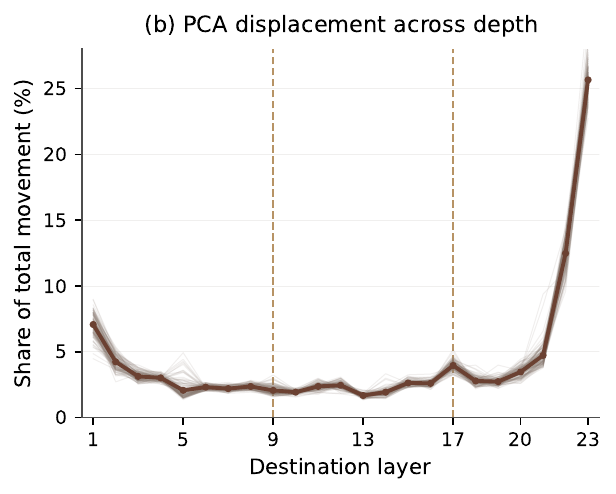}
    \hspace{0.015\textwidth}
    \includegraphics[width=0.32\textwidth]
    {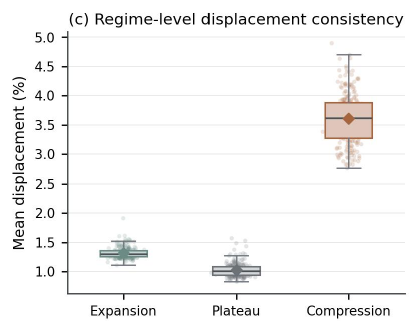}

    \caption{
    \textbf{Representation transformation dynamics across depth.}
    (a) Mean pairwise layer similarity, with linear CKA above the diagonal
    and cosine similarity below.
    (b) Adjacent-layer displacement measured in a shared PCA basis.
    (c) Dataset-level displacement across the expansion, plateau, and
    compression regimes.
    }
    \label{fig:regime_dynamics}
\end{figure}

The ED trajectory characterizes how broadly representation variance is
distributed across depth. To quantify the strength of transformation within
each regime, we complement ED with pairwise layer similarity, measured by
linear CKA \citep{kornblith2019similarity} and cosine similarity, together
with adjacent-layer displacement in a shared PCA space. Metric details are
provided in Appendix~\ref{app:pca_displacement}.

Figure~\ref{fig:regime_dynamics}(a) reveals distinct depthwise
transformation dynamics. During expansion, representations progressively
depart from earlier layers. The plateau forms a pronounced high-similarity
block across both nearby and distant layers, indicating relative mid-depth
stabilization. Near the output, this block breaks down as the final layers
increasingly diverge from the plateau representations.

Representation displacement exhibits the same structure
(Figure~\ref{fig:regime_dynamics}(b)). Layer-to-layer movement remains
substantial during expansion, reaches its minimum through the plateau, and
rises sharply again at late depth.The final layer transitions exhibit the largest concentration of PCA displacement, indicating intensified representational reorganization near the output. This concentration is consistent
across datasets, with displacement lowest during the plateau and highest in
the compression regime
(Figure~\ref{fig:regime_dynamics}(c)).

Late compression therefore coincides with a concentrated phase of geometric
reorganization near the output. Representation transformation intensifies
again after the mid-depth plateau, showing that substantial internal
reorganization persists into the final layers.

\subsection{Early Predictive Accessibility Precedes Candidate Resolution}
\label{sec:candidate_resolution}

To characterize the decision structure associated with this late
reorganization, we analyze the geometry among competing class candidates.
For each support class $c$, we construct a layer-specific prototype
$\mu_c^{(\ell)}$ and compute its cosine similarity with query $i$,
$
s_{i,c}^{(\ell)}
=
\cos(h_i^{(\ell)},\mu_c^{(\ell)}).
$
We quantify candidate competition by the margin between the two most similar
class prototypes,
$
\Delta_i^{(\ell)}
=
s_{i,(1)}^{(\ell)}-s_{i,(2)}^{(\ell)},
$
and use the rank of the correct-class prototype to measure alignment with
the target class. The analysis covers all 105 multiclass datasets from
OpenML-CC18 and TALENT. Full diagnostic details are provided in
Appendix~\ref{app:candidate_resolution}.

\begin{figure}[!ht]
    \centering
    \includegraphics[width=\textwidth]
    {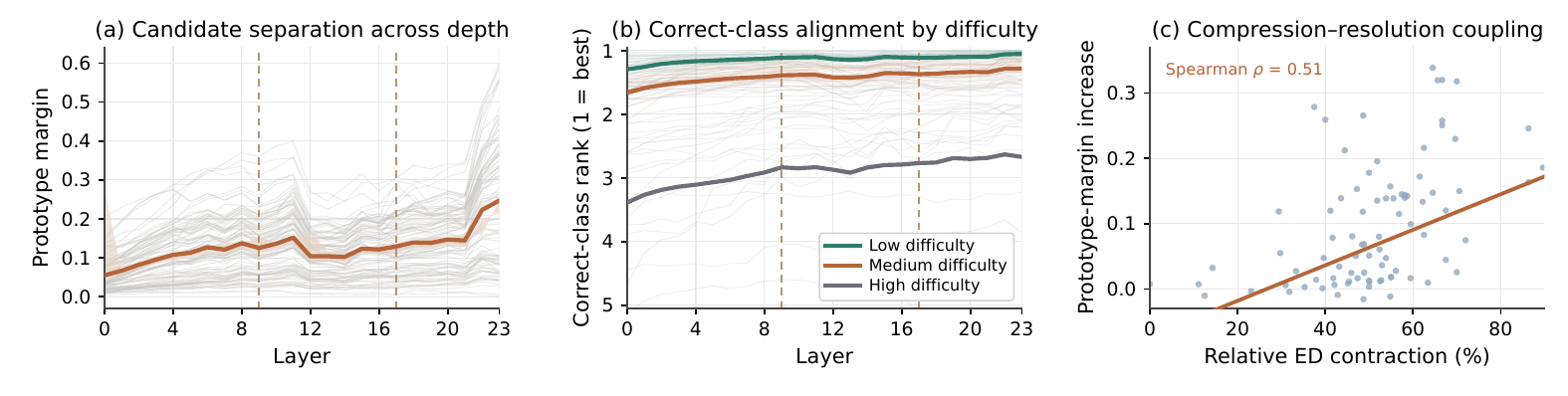}
    \caption{
    \textbf{Candidate separation strengthens at late depth.}
    (a) Prototype margins remain small through the early and middle layers
    and widen most strongly near the output.
    (b) Correct-class alignment emerges earlier on easier tasks, whereas
    harder tasks retain candidate ambiguity deeper into the network.
    (c) Stronger late effective-dimensionality contraction is associated
    with larger increases in prototype margin.
    }
    \label{fig:candidate_resolution}
    \vspace{-0.4\baselineskip}
\end{figure}

Figure~\ref{fig:candidate_resolution}(a) shows that prototype margins remain
small through the early and middle layers and widen most strongly near the
output. Figure~\ref{fig:candidate_resolution}(b) reveals a corresponding
progression in correct-class alignment: easier tasks establish favorable
alignment earlier, whereas harder tasks retain candidate ambiguity deeper
into the network.

Class-relevant information is already accessible well before strong candidate
separation emerges. The attn@V probe exhibits strong linear class decodability
by L2, whereas prototype margins widen primarily at later depth
(Appendix~\ref{app:attn_readability}). Predictive accessibility therefore
precedes strong geometric resolution among competing class candidates.

We next relate late candidate resolution to the compression identified in
Section~\ref{sec:dimensional_expansion}. For each dataset $d$, we define
relative late compression as
\[
\scalebox{0.80}{$\displaystyle
C_d =
\frac{\mathrm{ED}_d^{(20)}-\mathrm{ED}_d^{(23)}}
     {\mathrm{ED}_d^{(20)}}
$},
\]
and the corresponding increase in candidate separation as
$
R_d =
\Delta_d^{(23)}-\Delta_d^{(20)}.
$
Figure~\ref{fig:candidate_resolution}(c) shows that datasets with stronger
late compression exhibit larger increases in prototype margin. The same
association persists across alternative specifications
(Appendix~\ref{app:compression_resolution_coupling}). Late dimensional
contraction and candidate resolution therefore covary systematically across
datasets.

These results separate predictive accessibility from geometric
resolution. Class-relevant information becomes readable at shallow depth
while competing candidates remain weakly separated. Near the output,
concentrated reorganization and compression coincide with stronger candidate
separation and target-aligned geometry. Predictive accessibility
therefore precedes geometric resolution.

\section{Predictions Commit at Different Depths}

\subsection{Prediction Commitment Is Sample-Dependent}
\label{sec:sample_commitment}

Section~\ref{sec:candidate_resolution} shows that predictive accessibility
can precede geometric resolution. We next turn to the sample level and ask
when the predicted class itself becomes stable. For sample $i$, we denote its layerwise prediction trajectory as
$p_i = (\hat{y}_i^{(0)}, \ldots, \hat{y}_i^{(23)})$
and define the stabilization depth as the earliest layer after which the
predicted class remains identical to the final-layer prediction:
$s_i = \min\{\ell : \hat{y}_i^{(k)} = \hat{y}_i^{(23)},\, \forall k \ge \ell\}$. We use \emph{predictive commitment} to denote this stabilization of class
identity. The resulting stabilization depth $s_i$ is a retrospective
trajectory-level quantity. Confidence and representation geometry may
continue to evolve after commitment.

\begin{figure}[!ht]
    \centering
    \includegraphics[width=\textwidth]
    {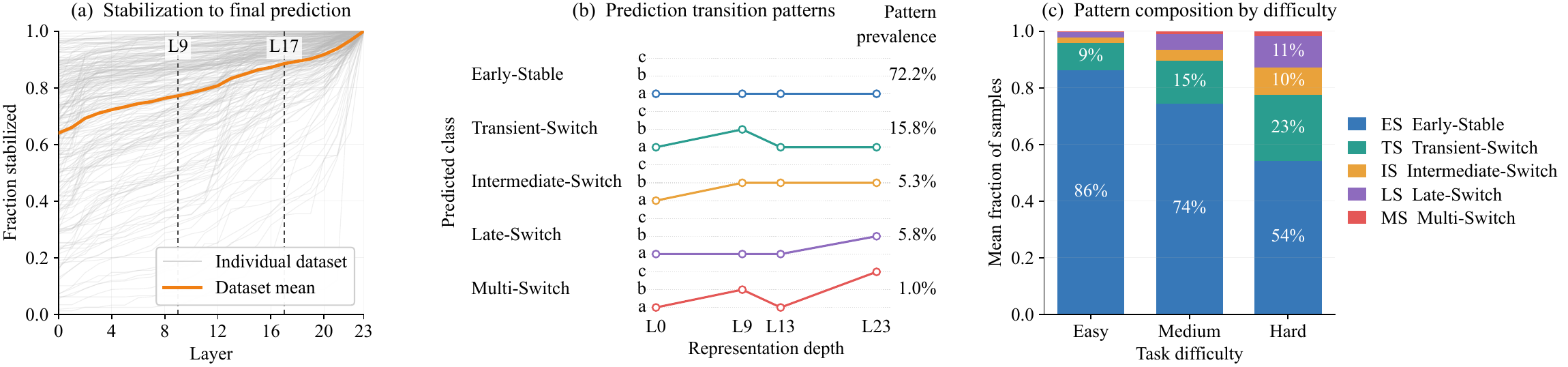}
    \caption{
    \textbf{Sample-level predictive commitment across depth.}
    (a) Cumulative fraction of samples stabilized to the final prediction
    across layers, shown per dataset and in aggregate.
    (b) Five recurring prediction-transition patterns and their pooled
    prevalence.
    (c) Pattern composition across tertiles of chance-normalized task
    difficulty; statistical analyses use difficulty as a continuous variable.
    }
    \label{fig:sample_dynamics}
\end{figure}

Figure~\ref{fig:sample_dynamics}(a) reveals substantial heterogeneity in
predictive commitment. Many predictions stabilize at shallow depth, while
others continue to change through intermediate or late layers, with pronounced
variation across datasets. Predictions therefore commit at markedly different
depths.

Stabilization depth captures \emph{when} a prediction becomes stable, while
the preceding trajectory reveals \emph{how} its class identity evolves before
commitment. We characterize this trajectory using prediction identities at
L9, L13, and L23 together with the total number of class transitions across
all 24 layers,
$n_{\mathrm{chg},i}
=
\sum_{\ell=1}^{23}
\mathbb{I}
\bigl[
\hat{y}_i^{(\ell)}
\neq
\hat{y}_i^{(\ell-1)}
\bigr]$ .
Pattern assignment uses only the prediction trajectory and does not depend on
ground-truth correctness. Full assignment rules are provided in
Appendix~\ref{app:trajectory_patterns}, with sensitivity to nearby
representative-layer choices reported in
Appendix~\ref{app:pattern_boundary_robustness}.

Figure~\ref{fig:sample_dynamics}(b) summarizes five recurring trajectory
patterns. \textbf{Early-Stable (ES)} predictions reach the final predicted
class early and remain there; \textbf{Transient-Switch (TS)} predictions
temporarily depart from it before returning; \textbf{Intermediate-Switch (IS)}
and \textbf{Late-Switch (LS)} predictions reach the final class at
intermediate and late depth, respectively; and \textbf{Multi-Switch (MS)}
predictions undergo repeated class changes. ES is dominant at 72.2\%, followed
by TS (15.8\%), LS (5.8\%), IS (5.3\%), and MS (1.0\%). Although early
commitment is predominant, more than one quarter of samples exhibit non-early
prediction dynamics. Similar pattern distributions are observed under
dataset-level aggregation and nearby representative-layer choices
(Appendices~\ref{app:pattern_aggregation}--\ref{app:pattern_boundary_robustness}).

Representation dynamics continue beyond early predictive commitment.
Early-Stable samples have stabilized by L9, while their representations still
exhibit the characteristic Expansion--Compression trajectory, including
pronounced late reorganization
(Appendix~\ref{app:es_post_commitment_geometry}). Predictive commitment can
therefore precede substantial later representational transformation. More
broadly, all samples traverse the same sequence of layers, while their
predictions commit at substantially different depths.

\subsection{Commitment Depth Tracks Task Difficulty and Local Ambiguity}
\label{sec:commitment_difficulty}

Commitment depth varies systematically with task difficulty and local
ambiguity. At the task level, we define model-relative, chance-normalized
difficulty as
\[
\scalebox{0.80}{$\displaystyle
D_d =
1 -
\frac{a_d - 1/C_d}{1 - 1/C_d}
$},
\]
where $a_d$ is final-layer accuracy and $C_d$ is the number of classes.
Harder tasks stabilize later on average, with a
consistent association across alternative definitions of difficulty
(Appendix~\ref{app:difficulty_robustness}). 

\begin{wrapfigure}{r}{0.45\textwidth}
    \vspace{-0.3\baselineskip}
    \centering
    \includegraphics[width=\linewidth]
    {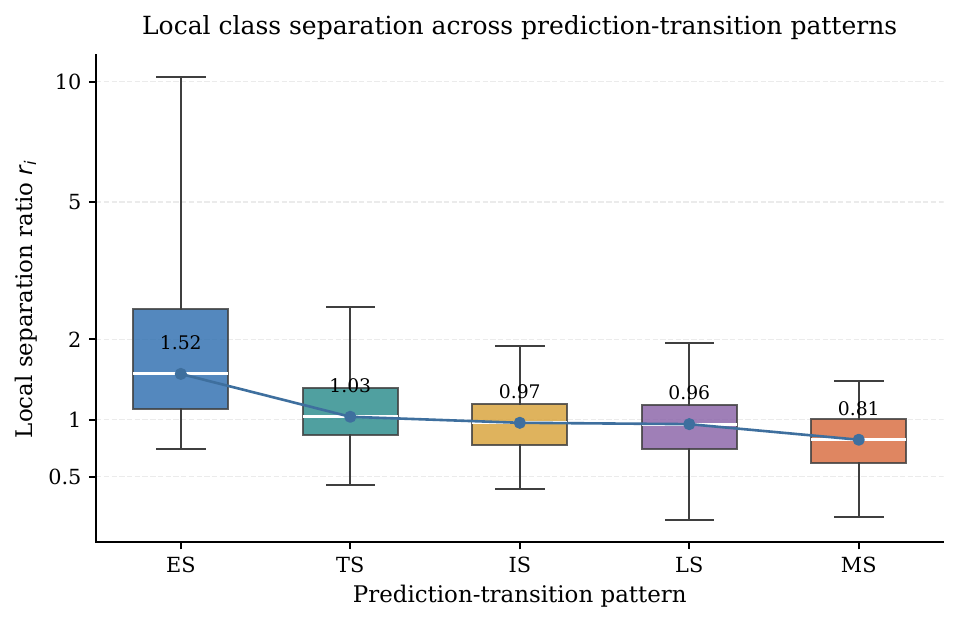}
    \caption{
    \textbf{Local class separation across prediction-transition patterns.}
    Separation decreases from ES to MS, indicating local ambiguity
    for more dynamic trajectories.
    }
    \label{fig:local_separation}
    \vspace{0.6\baselineskip}
\end{wrapfigure}

Correspondingly,
Figure~\ref{fig:sample_dynamics}(c) shows a systematic shift from ES toward
non-ES trajectories as task difficulty increases. At the sample level, we quantify local ambiguity using class separation in
the input space:
\[
\scalebox{0.80}{$\displaystyle
r_i =
\frac{
d\!\left(x_i,\mathrm{NN}_{\mathrm{diff}}\right)
}{
d\!\left(x_i,\mathrm{NN}_{\mathrm{same}}\right)
}
$},
\]
where $\mathrm{NN}_{\mathrm{same}}$ and $\mathrm{NN}_{\mathrm{diff}}$ denote
the nearest same- and different-class neighbors, respectively. Smaller $r_i$
indicates greater local ambiguity. Median separation decreases from $1.52$
for ES to $0.81$ for MS
(Figure~\ref{fig:local_separation}). Dataset-aware comparisons further show
higher separation for ES than for every non-ES pattern in approximately
95\% of shared datasets
(Appendix~\ref{app:local_separation_details}). For cross-dataset analyses, we convert $r_i$ to a within-dataset
local-separation percentile to place datasets with different raw distance
scales on a common reference. Lower percentiles correspond to more ambiguous
samples, whereas higher percentiles indicate more clearly separated local
neighborhoods.

Commitment depth therefore varies systematically across both tasks and
samples. Harder tasks commit later on average, while locally ambiguous samples
exhibit more dynamic prediction trajectories. Thus, inputs traversing the
same network can differ substantially in how much depth is required before
their predictions stabilize.

\section{The Functional Value of Depth}
\label{sec:functional_depth}

\subsection{Additional Depth Has Selective Utility}
\label{sec:late_utility}

We next examine whether the heterogeneous commitment identified above is
reflected in the functional value of additional computation at both the
sample and task levels. Lower local-separation percentiles indicate greater
ambiguity, whereas higher percentiles correspond to more clearly separated
samples.\begin{figure}[t]
    \centering
    \includegraphics[width=\textwidth]{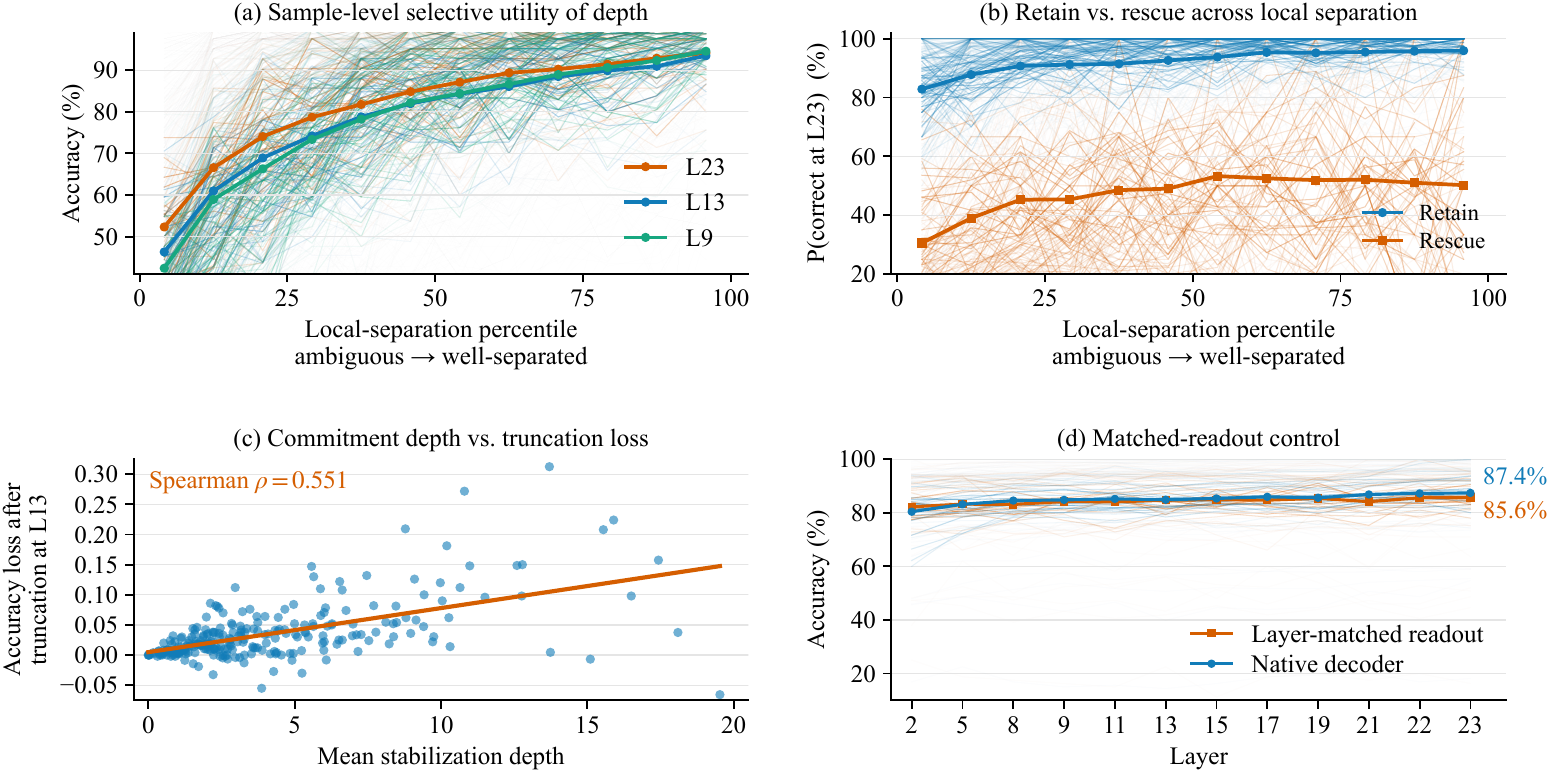}
    \caption{
    \textbf{Depth has selective utility across samples and tasks.}
   (a) Accuracy at L9, L13, and L23 across within-dataset local-separation
percentiles; locally ambiguous samples show larger gains with depth.
(b) Retention of L9-correct predictions and rescue of L9 errors across
local-separation percentiles; both increase with separation.
(c) Mean stabilization depth is positively associated with L13 truncation loss.
(d) Accuracy increases with depth under layer-matched readouts.
    }
    \label{fig:late_utility}
\end{figure} 

At the sample level, Figure~\ref{fig:late_utility}(a) shows that depthwise
benefit depends strongly on local ambiguity. Accuracy is already high at L9
for well-separated samples and changes only modestly toward L23, whereas the
gap between shallow and final depth grows as local separation decreases.
Locally ambiguous inputs therefore obtain larger aggregate gains from
additional computation. To distinguish aggregate depthwise gains from the fate of individual shallow
predictions, we separately measure retention of shallow successes and rescue
of shallow errors:
\[
\mathrm{Retain}
=
P(\hat{y}^{(23)}=y \mid \hat{y}^{(9)}=y),
\qquad
\mathrm{Rescue}
=
P(\hat{y}^{(23)}=y \mid \hat{y}^{(9)}\neq y).
\]
Figure~\ref{fig:late_utility}(b) shows that both retention and rescue increase
with local separation. Thus, ambiguous groups show larger aggregate gains
overall, even though errors already present at L9 are less frequently
rescued. The same trends hold with L8 or L10 as the shallow reference
(Appendix~\ref{app:shallow-anchor}).

The same selectivity appears at the task level. We define the loss from
truncating inference at layer $\ell$ as
$
\Delta \mathrm{Acc}^{(\ell)}_d
=
\mathrm{Acc}^{(23)}_d-\mathrm{Acc}^{(\ell)}_d .
$
Using L13 as the representative truncation point,
Figure~\ref{fig:late_utility}(c) shows that datasets with later mean
stabilization depth incur greater accuracy loss. The association remains
positive for later truncation points and with balanced accuracy
(Appendix~\ref{app:late_utility}). Later-committing tasks are therefore more
sensitive to truncation. The depthwise advantage also persists under depth-specific readouts.
With independently trained layer-matched readouts,
Figure~\ref{fig:late_utility}(d) shows that accuracy increases from
$82.2\%$ at L2 to $85.6\%$ at L23. Greater depth therefore remains
predictively beneficial when each layer is evaluated through its own
matched readout.

The functional value of depth is therefore selective: locally ambiguous
samples obtain larger aggregate gains from additional computation, while
later-committing tasks are more sensitive to truncation. Well-separated
inputs, by contrast, gain comparatively little from the remaining layers.

\subsection{Later Computation Is Ordered Rather Than Repetitive}
\label{sec:repeated_computation}
\label{sec:layer_order}

We next characterize the structure of this later computation through
interventions on the ICL stack, comparing repeated shared transformations with
the sequence of distinct learned blocks.

\begin{figure}[t]
    \centering
    \includegraphics[
        width=\linewidth,
        trim=0 40 0 0,
        clip
    ]{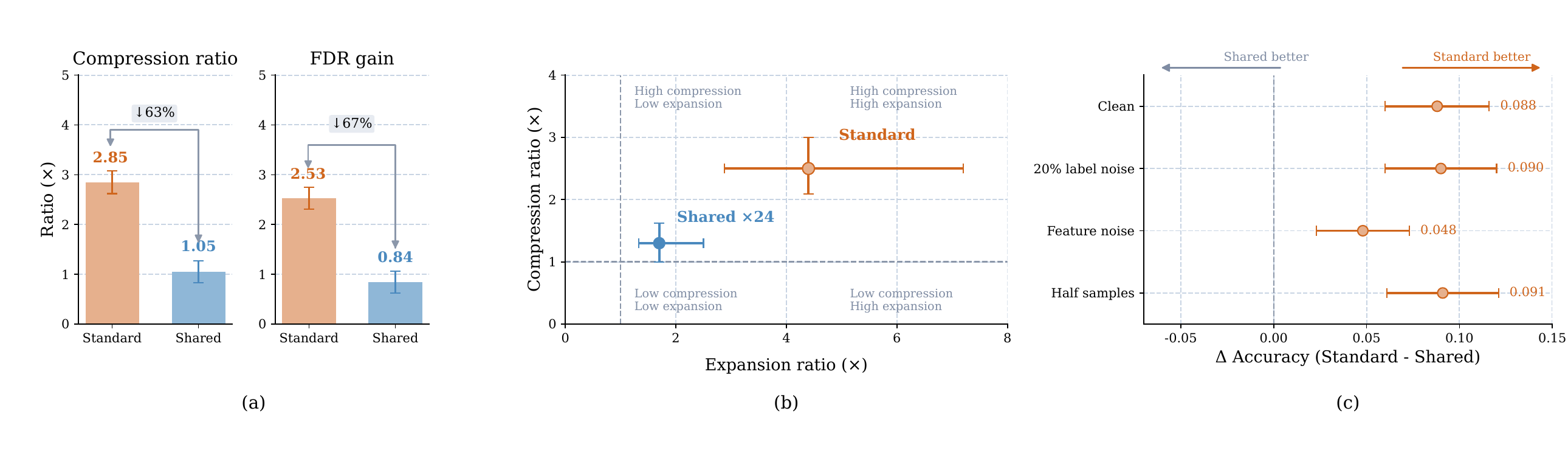}
    \caption{
    \textbf{Shared-weight repetition produces a distinct depthwise trajectory.}
    (a) Shared-weight repetition weakens compression and
    class-discriminative growth.
    (b) The repeated-block trajectory remains distinct at matched
    computational depth.
    (c) The full model retains a predictive advantage under perturbation.
    }
    \label{fig:depth_interventions}
\end{figure}

Replacing the sequence of distinct ICL blocks with repeated applications of
a shared block substantially weakens both late compression and
class-discriminative growth
(Figure~\ref{fig:depth_interventions}(a)). The effective-dimensional
compression ratio decreases from $2.85\times$ to $1.05\times$, while the FDR
gain decreases from $2.53$ to $0.84$. The repeated-block trajectory remains
qualitatively distinct even after 24 applications
(Figure~\ref{fig:depth_interventions}(b)), with the same pattern observed for
representative blocks selected from different stages
(Appendix~\ref{app:shared_weight_controls}).

These representational differences are accompanied by predictive differences.
Figure~\ref{fig:depth_interventions}(c) shows a positive average advantage for
the original model under clean evaluation, label noise, feature perturbation,
and reduced sample availability. Under $20\%$ label noise, this advantage
holds on every evaluated task
(Appendix~\ref{app:robustness_matrix}). The learned sequence of distinct
blocks therefore yields both the characteristic representation trajectory and
stronger predictive behavior than repeated generic computation.
\begin{wrapfigure}{r}{0.45\textwidth}
    \vspace{-0.8\baselineskip}
    \centering
    \includegraphics[
        width=\linewidth
    ]{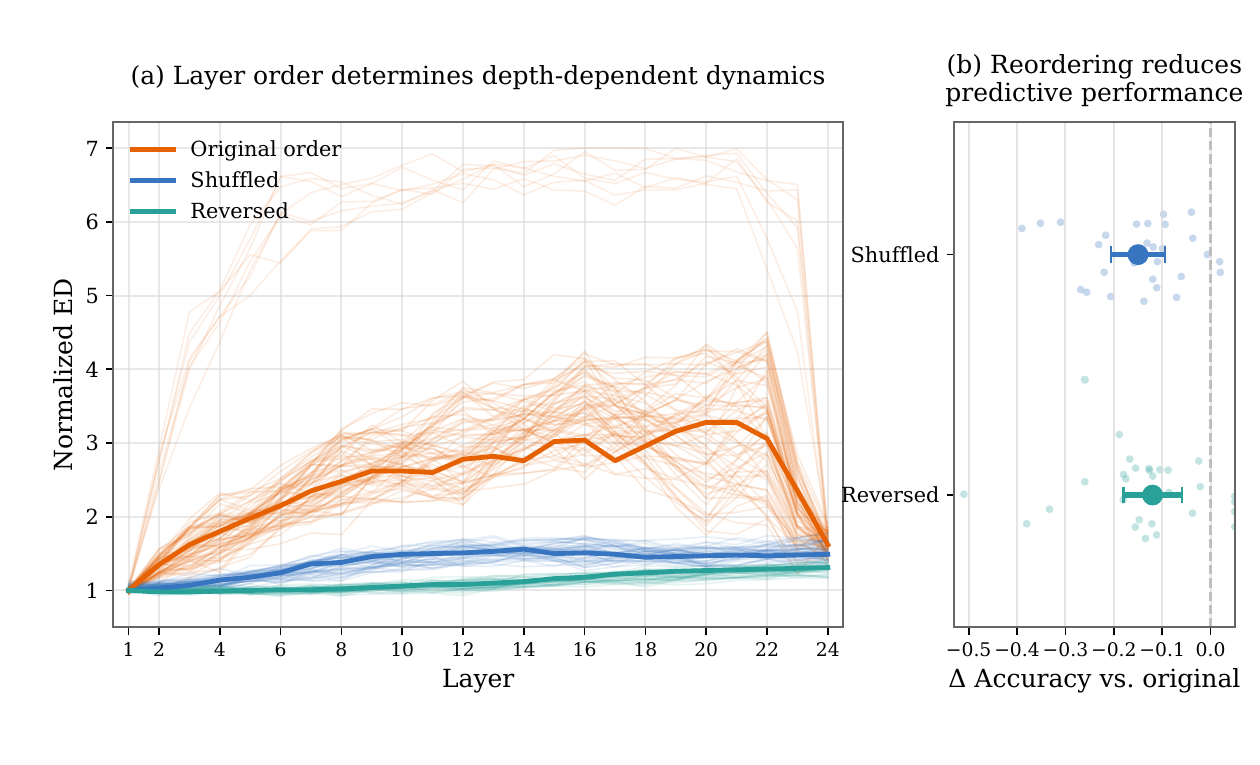}
    \caption{
    \textbf{Layer order is functionally consequential.}
    (a) Shuffling or reversing the learned sequence disrupts the
    depthwise representation dynamics.
    (b) Reordering reduces predictive performance on average.
    }
    \label{fig:layer_order}
    \vspace{-0.5\baselineskip}
\end{wrapfigure}
A complementary intervention preserves the learned blocks while changing only
their execution order. Shuffling or reversing the learned sequence largely removes the characteristic late compression in the representation trajectory (Figure~\ref{fig:layer_order}(a)), while also reducing predictive performance on most evaluated tasks (Figure~\ref{fig:layer_order}(b)). Because the learned transformations are
preserved, this intervention isolates execution order as a functional
component of depth.

Later computation is therefore structured as an ordered sequence of distinct learned operations. Its representational and predictive effects depend on both transformation identity and execution order. This computation path is shared across inputs, while their resolution requirements vary substantially.

\section{Conclusion}
\label{sec:discussion}

\textbf{Interpretation.}
Our study clarifies when later computation remains predictively useful and
how its value varies across inputs. Class-relevant information becomes
accessible at shallow depth, while representation geometry continues to evolve
through an \emph{Expansion--Compression} trajectory, with predictive
accessibility preceding geometric resolution. At the sample level, predictions
commit at substantially different depths, varying systematically with task
difficulty and local ambiguity. This heterogeneity is reflected in the
functional value of depth: locally ambiguous samples obtain larger aggregate
gains from additional computation, later-committing tasks are more sensitive
to truncation, and depthwise predictive gains remain under layer-matched
readouts. Structural interventions further show that later computation depends
on an ordered sequence of distinct learned transformations rather than generic
repetition. These findings show that TFMs follow a shared computation
path whose functional value varies substantially across inputs, motivating an
input-dependent view of depthwise redundancy.

\textbf{Limitations and Future Work.}
Our analysis focuses primarily on TabPFN-3 and classification, although
complementary experiments reveal related representation dynamics in several
additional TFMs. The strength and temporal organization of
\emph{Expansion--Compression} may vary across architectures and model scales.
A practical early-exit mechanism will require reliable online signals that
estimate whether additional computation is likely to alter or improve a
prediction, rather than relying on the retrospective stabilization depth used
in our analysis. Future work could investigate inference-time signals derived
from prediction confidence, trajectory consistency, or representation geometry,
and determine how reliably they identify inputs that benefit from continued
computation. Extending these analyses to broader TFM architectures and regression
settings is another important direction, particularly because regression will
require suitable continuous notions of predictive commitment rather than
discrete class-identity stabilization.

\section*{Reproducibility Statement}

We have made every effort to ensure that the results presented in this paper
are reproducible. The experimental setup, including the evaluated models,
datasets, layerwise evaluation protocols, representation metrics, prediction
analyses, and structural interventions, is described in detail in the main
paper and appendices. We additionally provide the definitions of all major
evaluation metrics, dataset-level aggregation procedures, statistical analyses,
and robustness checks used throughout the study. The datasets used in our experiments are drawn from publicly available
benchmark collections, including OpenML-CC18 and TALENT, enabling consistent
and reproducible evaluation. Detailed descriptions of model architectures and
layerwise representation extraction are provided in Appendix~A, while
Appendix~B specifies the representation metrics and analytical procedures.
Additional analyses of candidate geometry, prediction dynamics, truncation,
matched readouts, shared-weight controls, data perturbations, and layer-order
interventions are documented in the subsequent appendices. We believe these
details provide sufficient information to facilitate replication and
verification of our results.

\section*{AI Use Statement}

Large language models (LLMs) were used to assist with the writing and
polishing of the manuscript. Specifically, we used LLMs to refine language,
improve readability, enhance clarity, and support sentence rephrasing,
grammar checking, and the overall flow of the text. LLMs were also used for
literature retrieval and discovery to help identify relevant prior work and
reduce the risk of overlooking closely related studies. In addition, LLMs
were used to support research ideation, primarily to refine the research
questions. All AI-assisted outputs were reviewed and verified by the authors,
who take full responsibility for the final content of the work.
\bibliography{iclr2027_conference}
\bibliographystyle{iclr2027_conference}

\clearpage
\appendix

\section{Model Architectures and Layerwise Representation Extraction}
\label{app:experimental_setup}

This appendix provides the architectural and representation-extraction
details underlying our layerwise analyses. All empirical findings are
reported in the corresponding main-text and supplementary analyses.

\subsection{Model Architectures}
\label{app:model_architectures}

\paragraph{Primary architecture.}
We use TabPFN-3 as the primary model for layerwise analysis
\citep{grinsztajn2026tabpfn3}. The pretrained model remains frozen
throughout all representation and inference analyses.

TabPFN-3 maps heterogeneous tabular features into fixed-dimensional row
representations before performing in-context learning over rows. Its
architecture consists of three stages. Stage~1 performs column-wise
feature-distribution embedding through three induced-attention blocks.
Stage~2 aggregates feature information within each row using three
transformer blocks and four learned CLS tokens. The resulting four
128-dimensional vectors are flattened into a 512-dimensional row
representation. Stage~3 applies 24 in-context-learning transformer blocks.
All layerwise analyses in this paper concern these 24 Stage-3 blocks.

\begin{table}[!ht]
\centering
\caption{
\textbf{Architecture of TabPFN-3 used in the primary analysis.}
Layer indices L0--L23 refer exclusively to the 24 Stage-3 ICL blocks.
}
\label{tab:tabpfn3_architecture}
\small
\setlength{\tabcolsep}{4.5pt}
\begin{tabular}{lll}
\toprule
Component & Hyperparameter & Value \\
\midrule
\multicolumn{3}{l}{\textit{Stage 1: Feature-distribution embedding}} \\
& Base embedding dimension & 128 \\
& Features per circular-shift group & 3 \\
& Induced-attention blocks & 3 \\
& Attention heads & 8 \\
& Inducing points per column & 128 \\
\midrule
\multicolumn{3}{l}{\textit{Stage 2: Feature aggregation}} \\
& Transformer blocks & 3 \\
& Attention heads & 8 \\
& CLS tokens per row & 4 \\
& Positional encoding & RoPE \\
& RoPE base & $100{,}000$ \\
\midrule
\multicolumn{3}{l}{\textit{Stage 3: In-context learning}} \\
& Row representation dimension & 512 \\
& Transformer blocks & 24 \\
& Query attention heads & 8 \\
& KV heads for support rows & 8 \\
& KV heads for query rows & 1 \\
& Feed-forward expansion & $2\times$ \\
\midrule
\multicolumn{3}{l}{\textit{Classification decoder}} \\
& Decoder type & Attention retrieval \\
& Decoder heads & 6 \\
& Head dimension & 64 \\
& Pretraining class limit & 160 \\
\bottomrule
\end{tabular}
\end{table}

TabPFN-3 uses RMSNorm throughout the network. Its classification decoder
is an attention-based retrieval module: support representations serve as
keys, their one-hot labels as values, and query representations as
queries. The decoder operates on the 512-dimensional Stage-3
representations and provides the model-native predictive readout used in
our layerwise analyses.

\paragraph{Cross-model architectures.}
To examine cross-model generality, we additionally consider TabPFN-v2,
TabPFN-2.5, TabICL, and TabICLv2
\citep{hollmann2025tabpfn,grinsztajn2025tabpfn25,
qu2025tabicl,qu2026tabiclv2}.
Table~\ref{tab:model_family_summary} summarizes the architectural
properties most relevant to comparing their layerwise representations.
All models are evaluated using their pretrained parameters without
task-specific backbone fine-tuning.

\begin{table*}[t]
\centering
\caption{
\textbf{Architectures used in the cross-model analyses.}
``ICL depth'' denotes the sequence of transformer blocks over which
layerwise representations are analyzed. Parameter counts correspond to
the classification models where available.
}
\label{tab:model_family_summary}
\small
\setlength{\tabcolsep}{4.5pt}
\begin{tabular}{lcccccl}
\toprule
Model &
Parameters &
ICL depth &
ICL dim. &
Pre-ICL stages &
Normalization \\
\midrule

TabPFN-v2
& 7.24M & 12 & 192
& Alternating feature/item attention
& Post-LN \\

TabPFN-2.5
& 10.72M & 24 & 192
& Alternating feature/item attention
& Post-LN \\

TabPFN-3
& 53M & 24 & 512
& Column embedding + row aggregation
& RMSNorm \\

TabICL
& 27.05M & 12 & 512
& TFcol (3) + TFrow (3)
& Pre-LN \\

TabICLv2
& -- & 12 & 512
& TFcol (3) + TFrow (3)
& Pre-LN \\

\bottomrule
\end{tabular}
\end{table*}

TabPFN-v2 performs in-context computation through alternating attention
over features and items. TabPFN-2.5 retains this representation strategy
while increasing the transformer depth from 12 to 24 blocks. Their
component-level architectures are summarized in
Table~\ref{tab:tabpfn_v2_v25_params}.

\begin{table*}[t]
\centering
\caption{
\textbf{Component-level architecture of TabPFN-v2 and TabPFN-2.5.}
}
\label{tab:tabpfn_v2_v25_params}
\small
\setlength{\tabcolsep}{6pt}
\begin{tabular}{lrr}
\toprule
Component & TabPFN-v2 & TabPFN-2.5 \\
\midrule
Total parameters
    & $7{,}244{,}554$
    & $10{,}718{,}218$ \\
Feature encoder
    & 768
    & 1,152 \\
Label encoder
    & 576
    & 576 \\
Positional embedding
    & 9,408
    & 9,408 \\
Transformer depth
    & 12
    & 24 \\
Transformer-block parameters
    & $7{,}077{,}888$
    & $10{,}616{,}832$ \\
Parameters per block
    & 589,824
    & 442,368 \\
Feature-attention module
    & 147,456
    & 147,456 \\
Item-attention module
    & 147,456
    & 147,456 \\
MLP
    & $147{,}456 + 147{,}456$
    & $73{,}728 + 73{,}728$ \\
Decoder
    & $148{,}224 + 7{,}690$
    & $74{,}112 + 3{,}850$ \\
\bottomrule
\end{tabular}
\end{table*}

\paragraph{TabICL.}
TabICL separates feature encoding from dataset-level in-context learning.
A column-embedding module first processes individual features, followed
by a row-interaction module and a 12-layer ICL predictor.
Table~\ref{tab:tabicl_v1_params} summarizes its component-level
architecture.

\begin{table*}[t]
\centering
\caption{
\textbf{Component-level architecture of TabICL.}
}
\label{tab:tabicl_v1_params}
\small
\setlength{\tabcolsep}{5pt}
\begin{tabular}{llr}
\toprule
Stage & Component & Parameters \\
\midrule

\multirow{8}{*}{Column embedder}
& Total & 877,824 \\
& Input linear & 256 \\
& Transformer blocks (3) & 844,032 \\
& Each block & 281,344 \\
& Attention 1 & $66{,}048 + 16{,}512$ \\
& MLP 1 & $33{,}024 + 32{,}896$ \\
& Attention 2 & $132{,}480 + 66{,}048$ \\
& Output linear & $16{,}512 + 256$ \\
\midrule

\multirow{5}{*}{Row interaction}
& Total & 398,216 \\
& Rotary embedding & 8 \\
& Transformer blocks (3) & 397,440 \\
& Each block & 132,480 \\
& Attention & $66{,}048 + 16{,}512$ \\
\midrule

\multirow{5}{*}{ICL predictor}
& Total & $25{,}775{,}626$ \\
& Transformer blocks (12) & $25{,}233{,}408$ \\
& Each block & $2{,}102{,}784$ \\
& Attention & $1{,}050{,}624 + 262{,}656$ \\
& MLP & $525{,}312 + 524{,}800$ \\
\midrule

Prediction head
& Linear decoder & 535,562 \\
\midrule

\multicolumn{2}{l}{Total model parameters}
& $27{,}051{,}666$ \\
\bottomrule
\end{tabular}
\end{table*}

\paragraph{TabICLv2.}
TabICLv2 follows a related compression-then-ICL organization while
modifying feature grouping, target injection, and attention scaling.
Features are repeatedly grouped through circular shifts $(0,1,3)$,
followed by a column-wise Set Transformer
($\mathrm{TF}_{\rm col}$), a row-wise transformer
($\mathrm{TF}_{\rm row}$), and a dataset-level ICL transformer
($\mathrm{TF}_{\rm icl}$).
Table~\ref{tab:tabiclv2_config} summarizes the architectural configuration.

\begin{table}[t]
\centering
\caption{
\textbf{Architecture of TabICLv2 used in the cross-model analysis.}
}
\label{tab:tabiclv2_config}
\small
\setlength{\tabcolsep}{5pt}
\begin{tabular}{lll}
\toprule
Stage & Hyperparameter & Value \\
\midrule

\multirow{4}{*}{$\mathrm{TF}_{\rm col}$}
& Layers & 3 \\
& Inducing vectors & 128 \\
& Model dimension & 128 \\
& Attention heads & 8 \\
\midrule

\multirow{4}{*}{$\mathrm{TF}_{\rm row}$}
& Layers & 3 \\
& Model dimension & 128 \\
& Attention heads & 8 \\
& CLS tokens & 4 \\
\midrule

\multirow{3}{*}{$\mathrm{TF}_{\rm icl}$}
& Layers & 12 \\
& Model dimension & 512 \\
& Attention heads & 8 \\
\midrule

Target embedding
& $\mathrm{Embed}_{\rm TAE}$ & Lookup, 10 classes \\
& $\mathrm{Embed}_{\rm ICL}$ & Lookup, 10 classes \\
\midrule

Prediction head
& Hidden dimension & 1024 \\
& Output dimension & 10 \\
\midrule

Other
& LayerNorm & Pre-LN with bias \\
& FFN expansion & $2\times$ \\
& Activation & GELU \\
& QASSMax MLP hidden dim. & 64 \\
\bottomrule
\end{tabular}
\end{table}

The TabICLv2 paper does not report a single total parameter count for the
classification model. We therefore leave the parameter count unspecified
rather than infer an unreported value.

\subsection{Layerwise Representation Extraction}
\label{app:representation_extraction}

For TabPFN-3, we extract the output representation after each of the 24
Stage-3 ICL blocks. For a query set containing $n_q$ examples, the
post-block query representation matrix at depth $\ell$ is

\[
H_q^{(\ell)}
=
\left[
h_{q,1}^{(\ell)},\ldots,h_{q,n_q}^{(\ell)}
\right]^\top
\in \mathbb{R}^{n_q \times 512}.
\]

Each layer is represented by the row-level residual-stream output after
the corresponding ICL block. The preceding feature-distribution and
feature-aggregation stages are excluded from the depth index.
Consequently, L0 denotes the output of the first ICL block and L23
denotes the output of the twenty-fourth and final ICL block.

Support and query representations are retained at every depth. Query
representations form the primary object of the representation-geometry
and sample-level analyses. Support representations are used when required
for class-conditioned diagnostics and model-native decoding.

For the cross-model analyses, we analogously extract the post-block
representation after every block in each model's dataset-level ICL
sequence. Layer indices are therefore model-specific rather than matched
by absolute index across architectures.

\section{Definitions and Representation-Space Measurements}
\label{app:representation_measurements}

This appendix formalizes the representation-space quantities used throughout
our layerwise analyses. Unless otherwise specified, all measurements are
computed independently within each dataset using the same set of query
examples across layers. Cross-dataset aggregation is specified by the
corresponding analysis.

We characterize representation geometry through complementary quantities:
PCA-based effective dimensionality describes how representation variance is
distributed across principal directions; shared-PCA displacement measures
sample-wise movement across depth; linear CKA and cosine similarity quantify
pairwise representational similarity; and the Fisher discriminant ratio
measures label-conditioned class separation.

\subsection{Notation}
\label{app:representation_notation}

For a dataset containing $n$ query examples, let
\begin{equation}
H^{(\ell)}
=
\begin{bmatrix}
(h_1^{(\ell)})^\top \\
\vdots \\
(h_n^{(\ell)})^\top
\end{bmatrix}
\in \mathbb{R}^{n\times d}
\label{eq:representation_matrix}
\end{equation}
denote the query representation matrix after layer $\ell$, where
$h_i^{(\ell)}\in\mathbb{R}^{d}$ is the representation of query $i$.
For TabPFN-3, $\ell\in\{0,\ldots,23\}$ and $d=512$.

We define the sample-centered representation matrix as
\begin{equation}
\widetilde{H}^{(\ell)}
=
H^{(\ell)}
-
\mathbf{1}
\bigl(\bar{h}^{(\ell)}\bigr)^\top,
\qquad
\bar{h}^{(\ell)}
=
\frac{1}{n}
\sum_{i=1}^{n}
h_i^{(\ell)}.
\label{eq:centered_representation}
\end{equation}

All layerwise comparisons use matched query examples across depth.

\subsection{PCA-Based Effective Dimensionality}
\label{app:effective_dimensionality}

We quantify effective dimensionality using the number of principal
components required to explain a fixed fraction of the representation
variance. Let
\begin{equation}
\sigma_1^{(\ell)}
\ge
\sigma_2^{(\ell)}
\ge
\cdots
\label{eq:singular_values}
\end{equation}
denote the singular values of the centered representation matrix
$\widetilde{H}^{(\ell)}$.

The PCA-based effective dimensionality is
\begin{equation}
\operatorname{ED}\!\left(H^{(\ell)}\right)
=
\min
\left\{
k:
\frac{
\sum_{j=1}^{k}
\left(\sigma_j^{(\ell)}\right)^2
}{
\sum_j
\left(\sigma_j^{(\ell)}\right)^2
}
\ge 0.90
\right\}.
\label{eq:effective_dimensionality}
\end{equation}

A larger ED indicates that representation variance is distributed across
a broader set of principal directions, whereas a smaller ED indicates
greater concentration of variance in fewer directions. We use the term
\emph{PCA-based effective dimensionality} to emphasize that this quantity
is defined from the variance spectrum rather than from a nonlinear
manifold estimator.

The 90\% retained-variance threshold is used throughout the primary
analysis. Sensitivity to alternative thresholds is reported in the
robustness analyses.

\paragraph{Expansion and compression ratios.}
To summarize the non-monotonic effective-dimensionality trajectory within
each dataset, let
\begin{equation}
\ell^\ast
=
\arg\max_{\ell}
\operatorname{ED}\!\left(H^{(\ell)}\right)
\label{eq:ed_peak_layer}
\end{equation}
denote a layer at which effective dimensionality reaches its maximum.

We define the expansion ratio as
\begin{equation}
\rho_{\mathrm{exp}}
=
\frac{
\operatorname{ED}\!\left(H^{(\ell^\ast)}\right)
}{
\operatorname{ED}\!\left(H^{(0)}\right)
},
\label{eq:expansion_ratio}
\end{equation}
and the compression ratio as
\begin{equation}
\rho_{\mathrm{com}}
=
\frac{
\operatorname{ED}\!\left(H^{(\ell^\ast)}\right)
}{
\operatorname{ED}\!\left(H^{(L)}\right)
},
\label{eq:compression_ratio}
\end{equation}
where $L$ denotes the final analyzed layer, with $L=23$ for TabPFN-3.

Thus, $\rho_{\mathrm{exp}}$ measures the increase from the initial
representation to the maximum effective dimensionality reached across
depth, whereas $\rho_{\mathrm{com}}$ measures the subsequent contraction
from that maximum to the final representation. Both quantities are
computed from the complete layerwise trajectory within each dataset.

\subsection{Representation Displacement in a Shared PCA Space}
\label{app:pca_displacement}

Effective dimensionality characterizes the distribution of representation
variance but does not quantify how far individual sample representations
move across depth. We therefore measure adjacent-layer displacement in a
shared PCA coordinate system.

For each dataset, we concatenate the query representations from all
analyzed layers,
\begin{equation}
H_{\mathrm{joint}}
=
\begin{bmatrix}
H^{(0)} \\
H^{(1)} \\
\vdots \\
H^{(L)}
\end{bmatrix},
\label{eq:joint_representation}
\end{equation}
and fit PCA once to the resulting joint representation matrix.

We retain
\begin{equation}
k_{\mathrm{PCA}}
=
\min(10,d,n-1)
\label{eq:pca_dimension}
\end{equation}
principal directions. Let $P$ denote the resulting projection matrix and
$\mu_{\mathrm{joint}}$ the mean of $H_{\mathrm{joint}}$. The shared-PCA
representation of query $i$ at layer $\ell$ is
\begin{equation}
z_i^{(\ell)}
=
P^\top
\left(
h_i^{(\ell)}
-
\mu_{\mathrm{joint}}
\right)
\in
\mathbb{R}^{k_{\mathrm{PCA}}}.
\label{eq:shared_pca_projection}
\end{equation}

Adjacent-layer displacement is defined as
\begin{equation}
D_{\mathrm{PCA}}^{(\ell)}
=
\frac{1}{n}
\sum_{i=1}^{n}
\left\|
z_i^{(\ell+1)}
-
z_i^{(\ell)}
\right\|_2,
\qquad
\ell=0,\ldots,L-1.
\label{eq:pca_displacement}
\end{equation}

Using a single PCA basis places all layers in the same coordinate system
and makes adjacent-layer movement directly comparable across depth.

To quantify where representation movement is concentrated, for a set of
adjacent-layer transitions $\mathcal{I}$ we define
\begin{equation}
R_{\mathrm{disp}}(\mathcal{I})
=
\frac{
\sum_{\ell\in\mathcal{I}}
D_{\mathrm{PCA}}^{(\ell)}
}{
\sum_{\ell=0}^{L-1}
D_{\mathrm{PCA}}^{(\ell)}
}.
\label{eq:displacement_fraction}
\end{equation}

For the late-displacement statistic reported for TabPFN-3,
$\mathcal{I}$ contains the final six adjacent-layer transitions. This
quantity therefore measures the fraction of total depthwise representation
movement concentrated near the output.

\subsection{Pairwise Representation Similarity}
\label{app:pairwise_similarity}

We characterize the similarity between representations at arbitrary
layers $\ell$ and $m$ using two complementary measures: linear centered
kernel alignment and sample-wise cosine similarity.

\paragraph{Linear CKA.}
For centered representation matrices
$\widetilde{H}^{(\ell)}$ and $\widetilde{H}^{(m)}$, linear CKA is
\begin{equation}
\operatorname{CKA}
\left(
H^{(\ell)},H^{(m)}
\right)
=
\frac{
\left\|
\left(\widetilde{H}^{(\ell)}\right)^\top
\widetilde{H}^{(m)}
\right\|_F^2
}{
\left\|
\left(\widetilde{H}^{(\ell)}\right)^\top
\widetilde{H}^{(\ell)}
\right\|_F
\,
\left\|
\left(\widetilde{H}^{(m)}\right)^\top
\widetilde{H}^{(m)}
\right\|_F
}.
\label{eq:linear_cka}
\end{equation}

Larger CKA indicates greater similarity in the global linear
representational structure of the two layers.

\paragraph{Cosine similarity.}
For the same pair of layers, we compute the mean cosine similarity between
corresponding query representations,
\begin{equation}
S_{\mathrm{cos}}^{(\ell,m)}
=
\frac{1}{n}
\sum_{i=1}^{n}
\frac{
\left(h_i^{(\ell)}\right)^\top h_i^{(m)}
}{
\left\|h_i^{(\ell)}\right\|_2
\left\|h_i^{(m)}\right\|_2
}.
\label{eq:pairwise_cosine}
\end{equation}

This quantity measures sample-wise directional alignment independently of
vector magnitude. Final-layer alignment is obtained as the special case
$S_{\mathrm{cos}}^{(\ell,L)}$.

Linear CKA and cosine similarity are used jointly in the pairwise
layer-similarity matrices: CKA characterizes similarity in global
representational structure, while cosine similarity characterizes the
directional alignment of corresponding sample representations.

\subsection{Fisher Discriminant Ratio}
\label{app:fisher_ratio}

To quantify label-conditioned class separation, we compute the Fisher
discriminant ratio (FDR) from query representations and query labels.

For class $c$, let $n_c$ denote the number of query examples belonging to
that class, $\mu_c^{(\ell)}$ their mean representation at layer $\ell$,
and $\mu^{(\ell)}$ the global query mean.

The between-class scatter matrix is
\begin{equation}
S_b^{(\ell)}
=
\sum_c
n_c
\left(
\mu_c^{(\ell)}
-
\mu^{(\ell)}
\right)
\left(
\mu_c^{(\ell)}
-
\mu^{(\ell)}
\right)^\top,
\label{eq:between_class_scatter}
\end{equation}
and the within-class scatter matrix is
\begin{equation}
S_w^{(\ell)}
=
\sum_c
\sum_{i:y_i=c}
\left(
h_i^{(\ell)}
-
\mu_c^{(\ell)}
\right)
\left(
h_i^{(\ell)}
-
\mu_c^{(\ell)}
\right)^\top.
\label{eq:within_class_scatter}
\end{equation}

We define
\begin{equation}
\operatorname{FDR}^{(\ell)}
=
\frac{
\operatorname{tr}
\left(
S_b^{(\ell)}
\right)
}{
\operatorname{tr}
\left(
S_w^{(\ell)}
\right)
}.
\label{eq:fisher_discriminant_ratio}
\end{equation}

Larger FDR indicates greater between-class variation relative to
within-class variation and therefore stronger geometric class separation.
Query labels are used for this retrospective representation-space
diagnostic.

In the structural-intervention analyses, \emph{FDR gain} denotes the
ratio between final- and initial-layer FDR,
$\operatorname{FDR}^{(L)}/\operatorname{FDR}^{(0)}$.

\subsection{Representative Analytical Layers}
\label{app:representative_layers}

All primary representation-space measurements are computed across the
complete layerwise trajectory. For TabPFN-3, we use L9 and L17 as
descriptive anchors for regime-level summaries. The corresponding
transition intervals are L0--L9 for expansion, L9--L17 for the plateau,
and L17--L23 for compression. These intervals summarize the aggregate
effective-dimensionality trajectory rather than defining separate
dataset-specific change points.

The late interval contains the local rebound near L20 followed by the
pronounced contraction toward L23. Accordingly, L20 is used as a
representative late-depth anchor near the aggregate effective-dimensionality
maximum, while L23 denotes the final representation.

Several representative depths recur in subsequent analyses, as summarized
in Table~\ref{tab:representative_layers}.

\begin{table}[t]
\centering
\caption{
\textbf{Representative TabPFN-3 layers used in subsequent analyses.}
The Expansion--Compression trajectory is characterized from the complete
layerwise measurements; these depths serve as recurring analytical anchors.
}
\label{tab:representative_layers}
\small
\setlength{\tabcolsep}{5pt}
\begin{tabular}{ll}
\toprule
Layer & Role in subsequent analyses \\
\midrule
L9  & Early-depth anchor; end of the expansion summary interval \\
L13 & Intermediate plateau anchor \\
L17 & Late-depth anchor; start of the compression summary interval \\
L20 & Late-depth anchor near the aggregate ED maximum \\
L23 & Final-layer representation \\
\bottomrule
\end{tabular}
\end{table}

L9 and L13 are used as representative depths in the sample-level
prediction-trajectory analyses. L13, L17, and L20 are additionally used
as truncation points in the functional-depth experiments. Robustness to
nearby representative-layer choices is evaluated separately.

\section{Cross-Model Generality of Expansion--Compression}
\label{app:cross_model_generality}

The main analyses focus on TabPFN-3 because its 24-layer ICL stack
provides a particularly clear setting in which to study depth-dependent
representation dynamics. We next examine whether the broad
Expansion--Compression pattern also appears in other pretrained tabular
foundation models.

We consider four additional models: TabPFN-v2, TabPFN-2.5, TabICL, and
TabICLv2. TabPFN-3 is omitted from the figures in this appendix because
its representation trajectory is analyzed extensively in the main text.
The purpose of the present analysis is instead to examine whether related
geometric dynamics extend beyond the primary model.

Across the four auxiliary architectures, the cross-dataset mean
effective-dimensionality trajectories exhibit the same broad qualitative
ordering: effective dimensionality increases from its initial state,
reaches an intermediate maximum, and subsequently decreases. The magnitude
and temporal organization of this trajectory nevertheless differ
substantially across models. We therefore use Expansion--Compression to
describe this recurring representation-space pattern across the models
examined here, while allowing its strength and temporal organization to
remain architecture dependent.

\subsection{Cross-Model Evaluation Protocol}
\label{app:cross_model_protocol}

For the trajectory comparison, we restrict the analysis to classification datasets for which all four auxiliary models have valid
layerwise measurements. Within each dataset, the same evaluation examples
are used across the compared representation measurements.

Architectural details of the individual models are reported in
Appendix~\ref{app:model_architectures}, and all geometric quantities follow
the definitions in Appendix~\ref{app:representation_measurements}.

Because the models differ in both representation width and initial
effective dimensionality, raw ED values are not directly comparable across
architectures. For visualization, we therefore normalize effective
dimensionality separately within each dataset--model pair by its
first-layer value,

\begin{equation}
\widehat{\operatorname{ED}}^{(\ell)}
=
\frac{
\operatorname{ED}(H^{(\ell)})
}{
\operatorname{ED}(H^{(0)})
}.
\label{eq:app_crossmodel_normalized_ed}
\end{equation}

Thus, every normalized trajectory begins at one, and values above one
indicate expansion relative to that model's own initial representation.
This is not a min--max normalization. Consequently, architectures can
attain different normalized peak values when their representations undergo
different relative amounts of expansion.

The expansion and compression ratios reported below are computed from the
unnormalized within-dataset ED trajectories using the definitions in
Appendix~\ref{app:effective_dimensionality}. For each dataset--model pair,
the ratios are computed from its complete layerwise trajectory and are then
averaged acros datasets. Because the layer attaining
maximum ED can differ across datasets, the mean expansion ratio need not
equal the maximum of the cross-dataset mean normalized trajectory.

Final normalized ED is likewise computed within each dataset--model pair
before cross-dataset averaging. Expansion, compression, and final normalized
ED are therefore averaged as separate dataset-level quantities and need not
satisfy an exact algebraic relation after aggregation.

For PCA displacement, each adjacent-layer displacement is divided by the
total displacement of the corresponding dataset--model trajectory. This
allows the temporal concentration of representation movement to be compared
across architectures with different overall displacement scales.

The curves in Figure~\ref{fig:app_crossmodel_geometry} show the original
layerwise measurements together with a shape-preserving interpolation of
their cross-dataset means for visualization. The interpolation affects only
the rendering of the curves; all quantitative statistics are computed from
the original discrete layer values.

\subsection{Representation Trajectories Across Architectures}
\label{app:cross_model_trajectories}

\begin{figure*}[t]
    \centering
    \includegraphics[width=\textwidth]
    {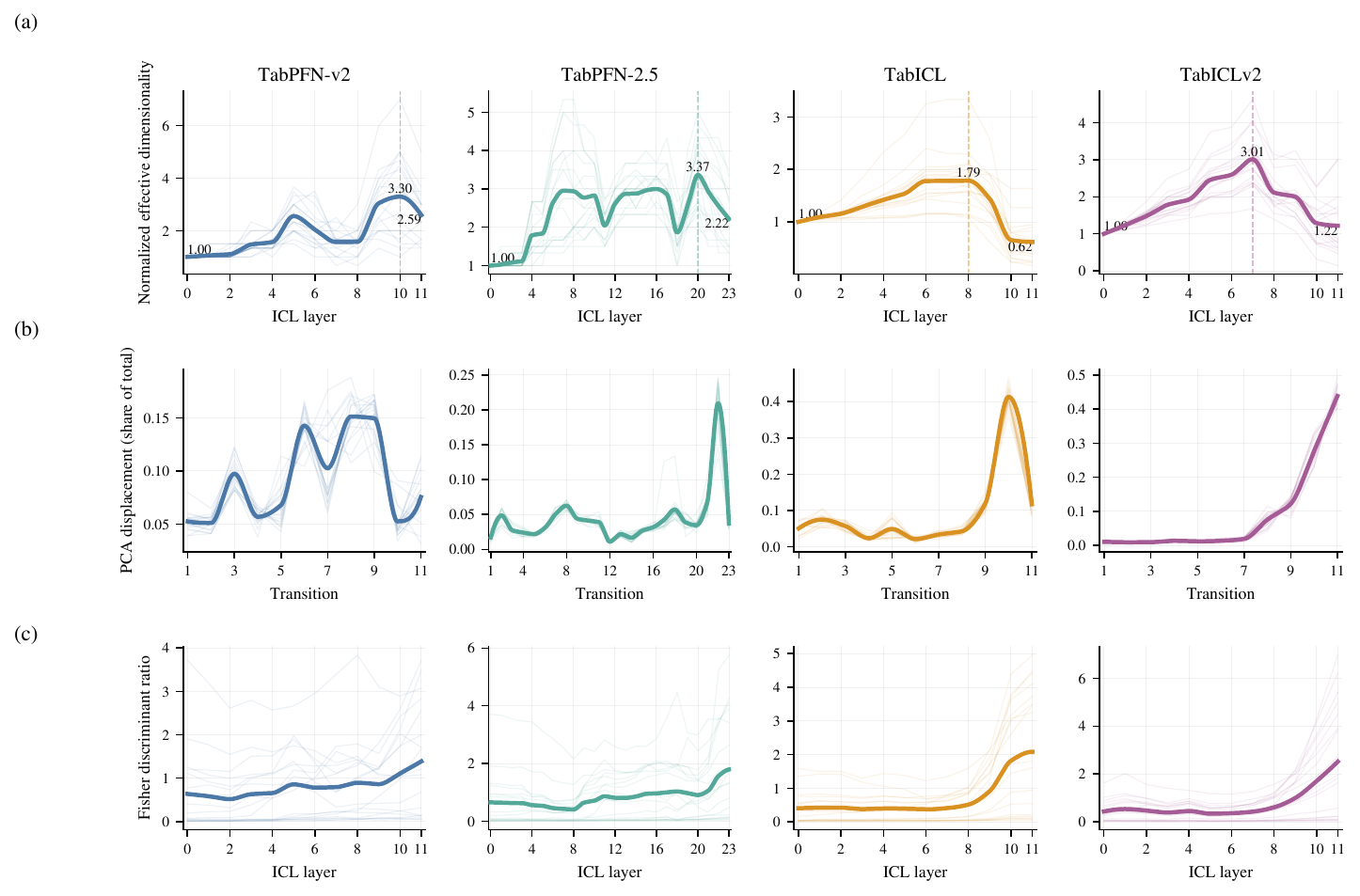}
    \caption{
    \textbf{Cross-model representation dynamics.}
    Results are shown for TabPFN-v2, TabPFN-2.5, TabICL, and TabICLv2 on
    matched datasets.
    \textbf{(a)} Effective dimensionality normalized by the first-layer
    value within each dataset--model pair.
    \textbf{(b)} Adjacent-layer PCA displacement expressed as a fraction
    of the total trajectory displacement.
    \textbf{(c)} Fisher discriminant ratio across depth.
    Thin curves denote individual datasets and thick curves denote the
    cross-dataset mean. Mean curves are interpolated only for visual
    smoothness; all statistics use the original layerwise measurements.
    Peak annotations in panel (a) refer to the cross-dataset mean
    trajectories and therefore need not equal the mean within-dataset
    expansion ratios reported in Table~\ref{tab:app_crossmodel_summary}.
    }
    \label{fig:app_crossmodel_geometry}
\end{figure*}

Figure~\ref{fig:app_crossmodel_geometry} compares effective dimensionality,
PCA displacement, and label-conditioned class separation across the four
auxiliary models.

The normalized ED trajectories show that non-monotonic
Expansion--Compression behavior is not restricted to TabPFN-3. All four
auxiliary models reach an intermediate representation with higher effective
dimensionality than their respective initial state.

The strength of expansion and subsequent compression nevertheless differs
markedly across architectures. Table~\ref{tab:app_crossmodel_summary}
summarizes these differences.

\begin{table}[t]
\centering
\caption{
\textbf{Summary of cross-model representation dynamics.}
Expansion and compression ratios are first computed within each dataset
and then averaged across matched datasets. Final ED denotes the
final effective dimensionality normalized by the first-layer value and is
likewise averaged from dataset-level values. Because these quantities are
averaged independently, their cross-dataset means need not satisfy exact
algebraic ratio identities. Late PCA reports the fraction of total PCA
displacement occurring in the final quarter of layer transitions.
}
\label{tab:app_crossmodel_summary}
\small
\setlength{\tabcolsep}{5pt}
\begin{tabular}{lccccc}
\toprule
Model
& Depth
& Expansion
& Compression
& Final ED
& Late PCA \\
\midrule
TabPFN-v2
& 12
& $3.61\times$
& $1.43\times$
& $2.59$
& $27.8\%$ \\

TabPFN-2.5
& 24
& $3.87\times$
& $1.87\times$
& $2.22$
& $44.4\%$ \\

TabICL
& 12
& $1.82\times$
& $3.88\times$
& $0.62$
& $64.9\%$ \\

TabICLv2
& 12
& $3.01\times$
& $3.59\times$
& $1.22$
& $83.8\%$ \\
\bottomrule
\end{tabular}
\end{table}

TabPFN-v2 and TabPFN-2.5 exhibit strong expansion but comparatively modest
peak-to-final contraction. Their final representations therefore remain
well above their initial effective dimensionality.

The two TabICL models exhibit a different organization. TabICL shows
weaker expansion but much stronger subsequent contraction, ending below
its initial normalized ED on average. TabICLv2 combines pronounced
expansion with similarly strong late contraction.

The PCA displacement trajectories reinforce this distinction. Only
$27.8\%$ of the total displacement in TabPFN-v2 occurs in the final
quarter of transitions, increasing to $44.4\%$ for TabPFN-2.5. The
corresponding fractions are substantially larger for TabICL
($64.9\%$) and TabICLv2 ($83.8\%$), indicating that representation
movement is considerably more concentrated near the end of their ICL
stacks.

The label-conditioned geometry follows a related but non-identical
trajectory. FDR generally increases toward later depth, indicating stronger
class separation in later representations, but the timing and magnitude of
this increase differ across architectures. ED, displacement, and
label-conditioned class separation therefore capture complementary aspects
of the depthwise representation trajectory.

\subsection{Pairwise Representation Similarity}
\label{app:cross_model_similarity}

The preceding analysis characterizes how individual geometric statistics
change with depth. We additionally examine the complete pairwise
relationship between representations from different layers.

\begin{figure*}[t]
    \centering
    \includegraphics[width=0.8\textwidth]
    {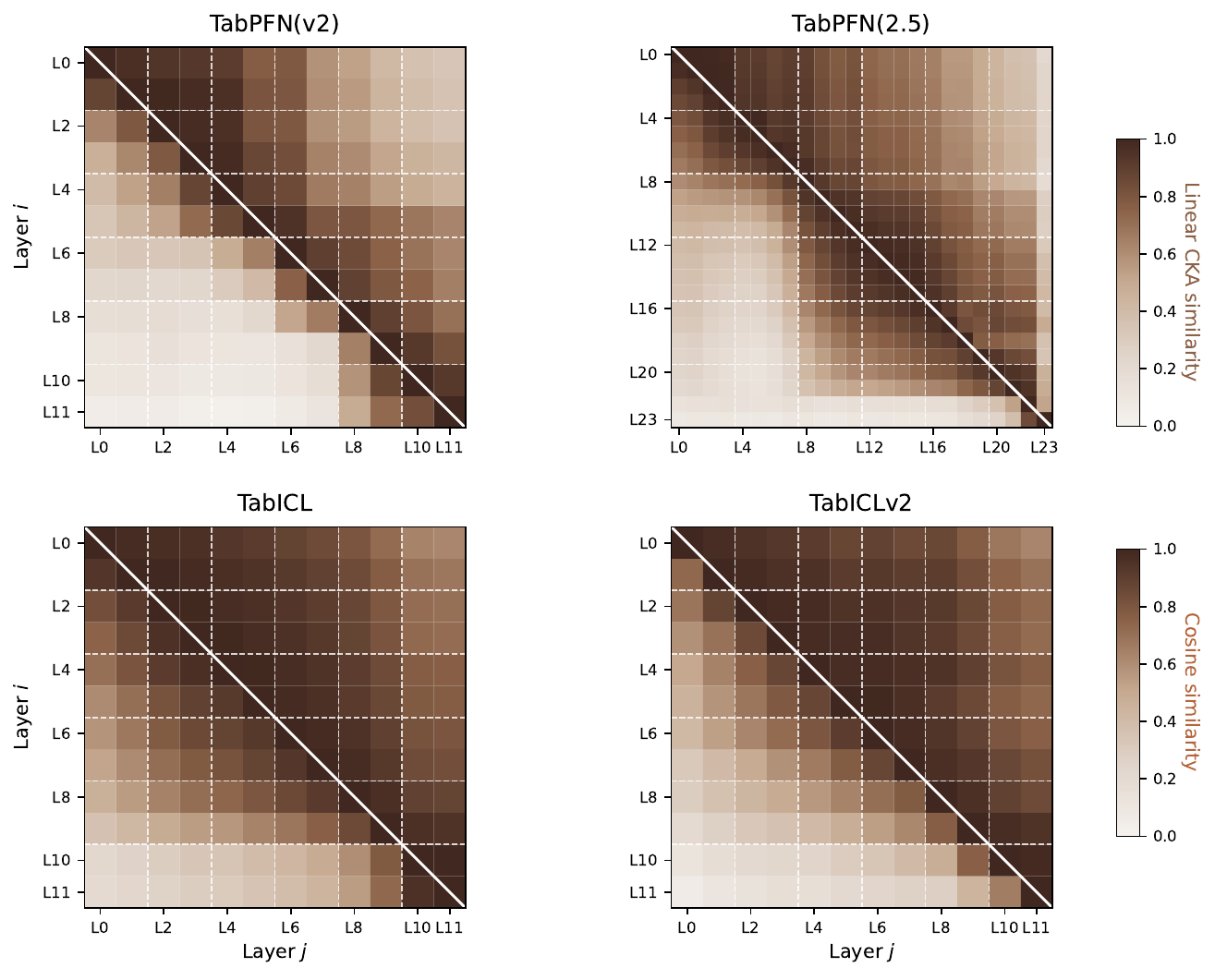}
    \caption{
    \textbf{Pairwise representation similarity across architectures.}
    For each model, the upper triangular matrix reports mean linear CKA
    between every pair of layers, while the lower triangular matrix reports
    mean cosine similarity. Darker entries indicate greater similarity.
    }
    \label{fig:app_crossmodel_similarity}
\end{figure*}

Figure~\ref{fig:app_crossmodel_similarity} combines two complementary
similarity measures. The upper triangular portion of each matrix reports
linear CKA, which measures similarity in global relational structure,
whereas the lower triangular portion reports mean cosine similarity,
which measures directional alignment of corresponding representation
vectors.

The similarity matrices reveal substantial architecture-dependent
structure that is not visible from ED alone. TabPFN-v2 and TabPFN-2.5
exhibit pronounced block-like regions in which nearby groups of layers
remain highly similar, separated by transitions with lower cross-block
similarity. This pattern is especially visible in the CKA matrices.

TabICL and TabICLv2 show broader regions of high CKA similarity across
depth. At the same time, their cosine similarity decreases more strongly
between temporally distant layers. Thus, two layers can preserve a
relatively similar global relational structure while their individual
representation vectors undergo substantial directional reorientation.

The distinction between CKA and cosine similarity is therefore informative
for interpreting representation change. High CKA can coexist with
substantial directional reorientation of individual vectors, while reduced
cosine similarity can occur without a comparable loss of global relational
structure. The two measurements provide complementary views of the same
depth-dependent trajectory.

\subsection{Scope of the Cross-Model Evidence}
\label{app:cross_model_scope}

Taken together, the cross-model results show that the
Expansion--Compression pattern observed in the primary TabPFN-3 analysis
is not unique to that architecture. Related non-monotonic ED trajectories
are also present in TabPFN-v2, TabPFN-2.5, TabICL, and TabICLv2.

At the same time, the models differ substantially in expansion strength,
compression strength, the temporal concentration of representation
movement, and the continuity of their pairwise representation structure.
The broad geometric pattern is therefore shared across the examined
architectures, whereas its detailed organization is model dependent.

The present comparison does not isolate a single architectural determinant
of these differences. Model depth, hidden dimension, attention
organization, decoder design, and pretraining procedure vary jointly across
the model families. Identifying which of these factors determines the
precise form of the trajectory therefore remains separate from the
cross-model generality established here.

Accordingly, the appropriate scope of the result is:
\emph{the Expansion--Compression phenomenon extends beyond TabPFN-3 within
the tabular foundation models examined here, while its strength and
temporal organization remain architecture dependent.}
\section{Architectural and Training Controls}
\label{app:architectural_training_controls}

To characterize the conditions under which Expansion--Compression emerges,
we conduct three complementary controls. We first randomize the learned
weights of TabPFN-3 while preserving its architecture. We then vary
computational depth within a lightweight TabPFN-style model while holding
the model family and training objective fixed. Finally, we compare the
resulting geometry with that of a conventionally supervised
FT-Transformer. Together, these experiments separate the roles of learned
parameterization, computational depth, and training paradigm in shaping the
observed representation trajectory.

\subsection{Randomized Weights Do Not Reproduce the Learned Trajectory}
\label{app:random_init_control}

We first isolate the contribution of learned parameterization in
TabPFN-3. We evaluate the same architecture under three weight conditions:
the original pretrained model, a model in which only the 24 ICL blocks are
randomized while the surrounding embedding and decoder components remain
pretrained, and a model in which all weights are randomized.

Table~\ref{tab:app_randomized_v3} reports the expansion and compression
ratios defined in Appendix~\ref{app:effective_dimensionality}.

\begin{table}[t]
\centering
\caption{
\textbf{Random-weight control for TabPFN-3.}
Each entry reports expansion ratio / compression ratio.
The pretrained model exhibits substantial Expansion--Compression, whereas
randomizing either the ICL stack or the complete model yields flat
effective-dimensionality trajectories on the tested datasets.
}
\label{tab:app_randomized_v3}
\small
\setlength{\tabcolsep}{5pt}
\begin{tabular}{lccc}
\toprule
Dataset
& Pretrained
& Randomized ICL blocks
& Fully randomized \\
\midrule
vehicle
& $4.00\times / 2.33\times$
& $1.00\times / 1.00\times$
& $1.00\times / 1.00\times$ \\

segment
& $4.67\times / 2.33\times$
& $1.00\times / 1.00\times$
& $1.00\times / 1.00\times$ \\

satimage
& $7.50\times / 3.33\times$
& $1.00\times / 1.00\times$
& $1.00\times / 1.00\times$ \\

diabetes
& $2.88\times / 2.09\times$
& $1.00\times / 1.00\times$
& $1.00\times / 1.00\times$ \\
\bottomrule
\end{tabular}
\end{table}

The pretrained network exhibits substantial expansion followed by
compression on every tested dataset. In contrast, both randomized
conditions yield expansion and compression ratios of one, corresponding to
flat effective-dimensionality trajectories.

Preserving the TabPFN-3 architecture while removing its learned
parameterization therefore eliminates the observed non-monotonic trajectory
in these controls. The Expansion--Compression profile is consequently tied
to the learned transformations of the pretrained model rather than to the
architecture alone.

\subsection{Controlled Depth in nanoTabPFN}
\label{app:nanotabpfn_depth}

The random-weight control establishes the importance of learned
transformations but does not isolate how computational depth contributes to
the trajectory. We therefore use nanoTabPFN, a lightweight trainable
TabPFN-style model for which variants with different numbers of transformer
layers can be trained under the same ICL objective.

We train variants with three, four, and five layers and examine their
layerwise effective-dimensionality trajectories on synthetic data drawn
from the training distribution. The model family, training objective, and
evaluation procedure are held fixed while depth is varied.

\begin{table}[t]
\centering
\caption{
\textbf{Controlled-depth experiment with nanoTabPFN.}
The three- and four-layer variants exhibit representation expansion without
a subsequent compression phase, whereas the five-layer variant exhibits
both expansion and compression.
}
\label{tab:app_nanotabpfn_depth}
\small
\setlength{\tabcolsep}{8pt}
\begin{tabular}{lccc}
\toprule
Depth
& ED trajectory
& Expansion
& Compression \\
\midrule
3 layers
& $1 \rightarrow 2 \rightarrow 2$
& $2.00\times$
& $1.00\times$ \\

4 layers
& $1 \rightarrow 1 \rightarrow 2 \rightarrow 2$
& $2.00\times$
& $1.00\times$ \\

5 layers
& $1 \rightarrow 2 \rightarrow 2 \rightarrow 2 \rightarrow 1$
& $2.00\times$
& $2.00\times$ \\
\bottomrule
\end{tabular}
\end{table}

Expansion is already present in the shallower variants: both the
three- and four-layer models reach twice their initial effective
dimensionality, but neither subsequently contracts. The five-layer model
instead follows
$1\rightarrow2\rightarrow2\rightarrow2\rightarrow1$,
returning to its initial effective dimensionality after reaching an
expanded intermediate state.

Within this controlled setting, increasing depth changes whether the
representation trajectory proceeds from expansion into compression.
Representation expansion therefore emerges at relatively shallow depth,
while additional computation can enable completion of the compression
phase.

\subsection{Supervised FT-Transformer Exhibits Contraction Without Pronounced Expansion}
\label{app:ft_transformer_control}

We finally examine whether the same geometric organization appears in a
conventionally supervised transformer for tabular data. We train an
FT-Transformer consisting of a feature tokenizer and eight transformer
encoder blocks with hidden dimension 64 and apply the same layerwise
effective-dimensionality analysis to CC18 classification datasets.

Across these datasets, the mean expansion ratio is only
$1.11\times$, whereas the mean compression ratio is
$2.33\times$. The dominant geometric trend is therefore contraction, with
only weak preceding expansion.

\begin{table}[t]
\centering
\caption{
\textbf{Representation geometry in the supervised FT-Transformer control.}
The supervised model exhibits substantial effective-dimensionality
contraction without the pronounced preceding expansion characteristic of
the Expansion--Compression trajectory.
}
\label{tab:app_ft_transformer_control}
\small
\setlength{\tabcolsep}{8pt}
\begin{tabular}{lc}
\toprule
Measurement
& FT-Transformer \\
\midrule
Mean expansion ratio
& $1.11\times$ \\

Mean compression ratio
& $2.33\times$ \\

Characteristic trajectory
& Predominantly contraction \\
\bottomrule
\end{tabular}
\end{table}

The supervised control therefore separates late contraction from the full
Expansion--Compression profile. Substantial contraction can arise under
conventional supervised transformer training, whereas the pronounced
preceding expansion observed in the pretrained TFMs is largely absent in
this control.

Taken together, the three controls delineate complementary properties of
the observed trajectory. Learned parameterization is required to reproduce
Expansion--Compression in TabPFN-3; additional depth enables the compression
phase to emerge in the controlled nanoTabPFN setting; and the supervised
FT-Transformer exhibits contraction without pronounced expansion. Within
the models examined here, these results support interpreting
Expansion--Compression as a learned, depth-dependent organization of ICL
representations rather than an automatic consequence of transformer depth.

\section{Methodological Validation and Robustness}
\label{app:methodological_validation}

This appendix provides additional methodological validation for our
layerwise analysis. We first establish why the native TabPFN-3 decoder can
be used as a common readout of intermediate representations, and then
separate native-decoder alignment from predictive information availability
using layer-matched and decoder-independent probes. Finally, we report
robustness checks for the geometric quantities underlying the Expansion--Compression analysis.

\subsection{Validating the Native TabPFN-3 Decoder as a Layerwise Readout}
\label{app:native_decoder}

\paragraph{Native many-class decoder.}
TabPFN-3 replaces the fixed-width classification head used in earlier
TabPFN variants with an attention-based retrieval decoder
\citep{grinsztajn2026tabpfn3}. For classification, training-set
representations act as keys, their corresponding one-hot labels act as
values, and query representations act as queries. Prediction is therefore
implemented as a learned soft nearest-neighbor retrieval over the
in-context support set.

Let $h_{m,\mathrm{q}}^{(\ell)} \in \mathbb{R}^{512}$ denote the
representation of query sample $m$ after ICL layer $\ell$, and let
$h_{n,\mathrm{s}}^{(\ell)} \in \mathbb{R}^{512}$ denote the corresponding
representation of support sample $n$. For decoder head $r$, the projected
query and key representations are

\begin{align}
q_{m}^{(\ell,r)}
    &= W_{Q}^{(r)} h_{m,\mathrm{q}}^{(\ell)}, \\
k_{n}^{(\ell,r)}
    &= W_{K}^{(r)} h_{n,\mathrm{s}}^{(\ell)} .
\end{align}

For notational simplicity, we write the attention weights using the
standard scaled dot-product form,

\begin{equation}
\alpha_{m,n}^{(\ell,r)}
=
\frac{
\exp\left(
{q_{m}^{(\ell,r)}}^{\top} k_{n}^{(\ell,r)}
/\sqrt{d_h}
\right)}
{
\sum_{j=1}^{N_{\mathrm{s}}}
\exp\left(
{q_{m}^{(\ell,r)}}^{\top} k_{j}^{(\ell,r)}
/\sqrt{d_h}
\right)
},
\end{equation}

where the actual TabPFN-3 implementation additionally applies its pretrained
QASSMax query scaling \citep{grinsztajn2026tabpfn3}. For class $c$, the
decoder aggregates the attention mass assigned to support samples of that
class and averages it across heads,

\begin{equation}
p_{m,c}^{(\ell)}
=
\frac{1}{H}
\sum_{r=1}^{H}
\sum_{n:\,y_n=c}
\alpha_{m,n}^{(\ell,r)} .
\end{equation}

The class logits are then

\begin{equation}
z_{m,c}^{(\ell)}
=
\log\!\left(
\operatorname{clip}\!\left(p_{m,c}^{(\ell)}\right)
\right).
\end{equation}

This is equivalent to the report's description of a head-averaged,
attention-weighted average of one-hot labels, written explicitly in
class-wise form. TabPFN-3 uses $H=6$ decoder heads with head dimension
$d_h=64$.

\paragraph{Why the decoder is well-defined across depth.}
All 24 ICL layers preserve the same 512-dimensional hidden representation,
and the decoder applies the same learned query and key projections
independently of layer index. The native decoder is therefore dimensionally
well-defined at every ICL layer and provides a fixed, model-native readout
across depth. This does \emph{not} imply a fixed decision boundary: both
query and support representations evolve with $\ell$, changing their
relative geometry and the resulting attention weights.

\paragraph{Decoder-alignment concern.}
Dimensional compatibility does not imply that native-decoder performance
measures all predictive information in an intermediate representation.
Earlier layers may contain label-relevant information while remaining
imperfectly aligned with the final decoder, motivating tuned or
layer-specific readouts \citep{belrose2023tuned,balef2026one}. We therefore
distinguish \emph{predictive decodability} from
\emph{native-decoder alignment}.

\subsubsection{Layer-Matched Readout Control}
\label{app:matched_readout}

To control for decoder mismatch, we train an independent linear readout for
each analyzed layer. Only query hidden states are used for this experiment:
within each dataset, query examples are split 50/50 into probe-training and
probe-evaluation subsets, and a separate classifier is fitted to
$h_{\mathrm{q}}^{(\ell)}$ for every layer $\ell$. This procedure removes
the requirement that an intermediate representation be aligned with the
native decoder while preserving the amount of predictive information
available at that layer.

\begin{figure}[t]
    \centering

    \begin{minipage}[t]{0.49\linewidth}
        \centering
        \includegraphics[width=\linewidth]
        {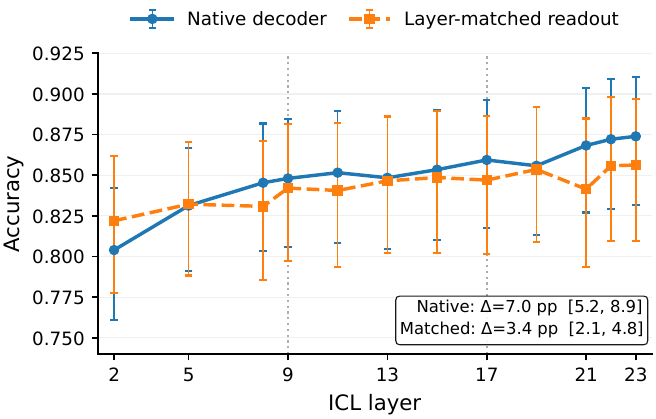}
    \end{minipage}
    \hfill
    \begin{minipage}[t]{0.49\linewidth}
        \centering
        \includegraphics[width=\linewidth]
        {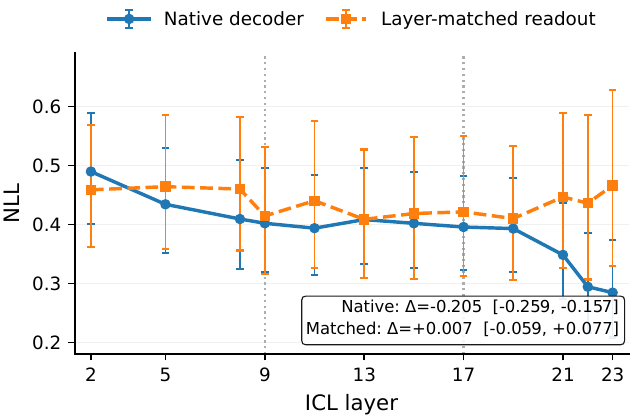}
    \end{minipage}

    \caption{
    \textbf{Native-decoder and layer-matched readout performance across
    depth.}
    \textbf{Left:} classification accuracy averaged over 72 OpenML-CC18
    tasks.
    \textbf{Right:} negative log-likelihood.
    Error bars denote 95\% bootstrap confidence intervals over datasets.
    Vertical dotted lines at L9 and L17 indicate representative analytical
    boundaries rather than estimated change points.
    }
    \label{fig:native_matched_readout}
\end{figure}

We evaluate both the native decoder and the layer-matched readout on all
OpenML-CC18 classification tasks. Figure~\ref{fig:native_matched_readout}
shows the dataset-level mean accuracy and negative log-likelihood (NLL);
error bars denote 95\% bootstrap confidence intervals obtained by resampling
datasets.

\paragraph{Accuracy.}
At L2, the native decoder reaches $0.804$ accuracy and the layer-matched
readout reaches $0.822$, indicating partial early-layer decoder
misalignment. From L2 to L23, matched-readout accuracy increases from
$0.822$ to $0.856$ ($+3.4$ pp; 95\% CI $[+2.1,+4.8]$ pp), while native
accuracy increases from $0.804$ to $0.874$ ($+7.0$ pp; 95\% CI
$[+5.2,+8.9]$ pp). Thus, the overall depthwise accuracy increase is not
explained by output-head mismatch alone.

\paragraph{Probability quality.}
Native-decoder NLL decreases from $0.490$ at L2 to $0.284$ at L23
($\Delta=-0.205$; 95\% CI $[-0.259,-0.157]$), whereas the matched-readout
change is small and statistically unresolved
($\Delta=+0.007$; 95\% CI $[-0.059,+0.077]$). We use NLL as a measure of
probabilistic prediction quality rather than as a standalone measure of
calibration.

\subsection{Decoder-Independent Probe Control}
\label{app:independent_probe}

The matched-readout experiment controls specifically for alignment with
TabPFN-3's final decoder. We additionally perform a broader
decoder-independent probing analysis to test whether the representation
dynamics identified in the main paper persist without using the native
decoder at all.

Following the general probing setup used in prior TFM analyses
\citep{balef2026one}, we extract query hidden states from each layer and
train both a linear classifier and a capacity-controlled MLP probe with
hidden dimension 64. Because support rows explicitly incorporate
in-context labels, only query representations are used as probe examples.
For each dataset, query examples are divided equally into probe-training and
probe-evaluation subsets. We report accuracy, NLL, and ECE. The experiment
is run on all 72 OpenML-CC18 tasks for TabPFN-3, TabICL, and TabICLv2. Across the 72 tasks, all three tested models exhibit higher MLP-probe
accuracy in later than in earlier layers:

\begin{figure}[t]
    \centering
    \includegraphics[
        width=\linewidth,
        trim=3cm 6cm 4cm 8cm,
        clip
    ]{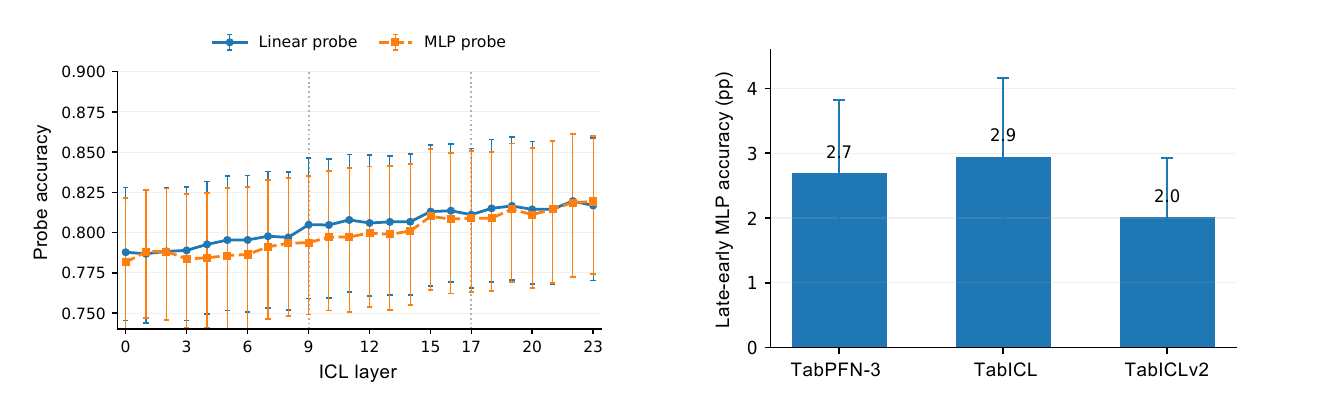}
    \caption{
    \textbf{Decoder-independent predictive decodability across depth.}
    \textbf{(a)} Linear and capacity-controlled MLP probe accuracy across
    the 24 ICL layers of TabPFN-3, averaged over 72 OpenML-CC18 tasks.
    Error bars denote 95\% bootstrap confidence intervals over datasets;
    dotted vertical lines indicate the representative L9 and L17 analytical
    boundaries.
    \textbf{(b)} Change in MLP-probe accuracy from the first to the final
    third of the ICL layers for TabPFN-3, TabICL, and TabICLv2.
    Differences are computed within each dataset before aggregation, and
    error bars denote 95\% bootstrap confidence intervals over the 72 tasks.
    }
    \label{fig:independent_probe}
\end{figure}

Figure~\ref{fig:independent_probe} shows an overall increase in
decoder-independent predictive accuracy across depth in TabPFN-3, despite
local fluctuations, together with positive late-over-early MLP differences
for TabPFN-3, TabICL, and TabICLv2.

\FloatBarrier

Table~\ref{tab:independent_probe} summarizes the corresponding
cross-model decoder-independent probe results.

\begin{table}[t]
\centering
\caption{
Decoder-independent probing across the 72 OpenML-CC18 tasks.
$\Delta$ denotes the mean difference between the final and initial thirds
of the layer sequence. ``Positive'' denotes the fraction of datasets with a
positive MLP-accuracy difference.
}
\label{tab:independent_probe}
\vspace{4mm}
\small
\begin{tabular}{lcccc}
\toprule
Model
& $\Delta$ MLP Acc.
& Positive
& $\Delta$ ECE
& $\Delta$ MLP NLL \\
\midrule
TabPFN-3  & +0.027 & 64\% & -0.005 & -0.056 \\
TabICL    & +0.029 & 66\% & -0.017 & -0.179 \\
TabICLv2  & +0.020 & 59\% & -0.008 & -0.104 \\
\bottomrule
\end{tabular}
\end{table}

For TabPFN-3, a phase-aligned analysis for
the detailed geometric study further separates changes in effective
dimensionality from changes in predictive decodability, as summarized in
Table~\ref{tab:probe_phase}:

\begin{table}[t]
\centering
\caption{
Phase-wise changes in effective dimensionality and decoder-independent
probe performance for TabPFN-3. The phase boundaries are representative
analytical boundaries derived from the aggregate ED trajectory.
}
\label{tab:probe_phase}
\vspace{4mm}
\small
\begin{tabular}{lrrrr}
\toprule
Phase
& $\Delta$ED
& $\Delta$ Linear Acc.
& $\Delta$ MLP Acc.
& $\Delta$ MLP NLL \\
\midrule
Expansion
& +10.95
& +0.0274
& +0.0278
& -0.291 \\
Plateau
& +0.95
& +0.0093
& +0.0150
& -0.011 \\
Compression
& -10.55
& -0.0040
& +0.0045
& +0.023 \\
Full depth
& +1.40
& +0.0327
& +0.0473
& -0.279 \\
\bottomrule
\end{tabular}
\end{table}

\FloatBarrier

Across the full depth of TabPFN-3, MLP-probe accuracy increases by
$4.73$ percentage points
($95\%~\mathrm{CI}=[+2.1,+8.1]$ pp), while MLP-probe NLL decreases by
$0.279$ ($95\%~\mathrm{CI}=[-0.462,-0.144]$). ED, by contrast, returns
close to its initial magnitude after expanding and then compressing.

Within the compression regime, the MLP-accuracy change is small
($+0.45$ pp) and its bootstrap interval includes zero. Thus, the control
does not establish a significant compression-phase increase in decodability;
it shows that the large reduction in ED occurs without a corresponding loss
of decoder-independent predictive performance.

\subsection{Robustness of the Geometric Analysis}
\label{app:geometry_robustness}

We test the geometric analysis against ED-threshold choice,
representation normalization, random-weight controls, representative phase
boundaries, and PCA coordinate artifacts. For a layerwise ED trajectory, we
define the expansion and compression ratios as
\begin{equation}
\rho_{\mathrm{exp}}
=
\frac{\max_{\ell}\mathrm{ED}(H^{(\ell)})}
     {\mathrm{ED}(H^{(0)})},
\qquad
\rho_{\mathrm{com}}
=
\frac{\max_{\ell}\mathrm{ED}(H^{(\ell)})}
     {\mathrm{ED}(H^{(L)})}.
\end{equation}
Values substantially above one indicate expansion from the initial
representation and subsequent compression toward the final representation,
respectively.

\paragraph{Sensitivity to the ED threshold.}

Our primary definition of ED uses the minimum number of principal
components required to explain 90\% of the variance. To test whether the
resulting trajectory depends on this choice, we recompute ED using variance
thresholds of 80\%, 85\%, 90\%, and 95\% on OpenML-CC18 datasets.
Table~\ref{tab:ed_threshold_sensitivity} reports the resulting expansion
and compression ratios.

\begin{table}[t]
\centering
\caption{
\textbf{Sensitivity to the effective-dimensionality threshold.}
Expansion and compression remain pronounced across variance thresholds from
80\% to 95\%. The last two columns report the fraction of datasets satisfying
$\rho_{\mathrm{exp}}>1.5$ and $\rho_{\mathrm{com}}>1.3$, respectively.
}
\label{tab:ed_threshold_sensitivity}
\vspace{2mm}
\small
\begin{tabular}{ccccc}
\toprule
Threshold
& $\rho_{\mathrm{exp}}$
& $\rho_{\mathrm{com}}$
& Exp. $>1.5$
& Comp. $>1.3$ \\
\midrule
80\% & 2.41$\times$ & 2.80$\times$ & 95\%  & 100\% \\
85\% & 2.40$\times$ & 2.57$\times$ & 95\%  & 100\% \\
90\% & 2.45$\times$ & 2.24$\times$ & 95\%  & 95\%  \\
95\% & 2.54$\times$ & 1.84$\times$ & 100\% & 100\% \\
\bottomrule
\end{tabular}
\end{table}

Across all four thresholds, the mean expansion ratio remains at least
$2.40\times$ and the mean compression ratio at least $1.84\times$; at
least 95\% of datasets satisfy each corresponding criterion.

\paragraph{Normalization control.}

Representation norms vary substantially across depth, raising the
possibility that changes in ED could be induced by scale variation rather
than by a change in the spectrum of representational degrees of freedom.
We therefore recompute the complete ED trajectory after three transformations:
row normalization, feature-wise standardization, and whitening.
The experiment is conducted on six datasets using the same layerwise
analysis pipeline. Table~\ref{tab:ed_normalization_control} summarizes
the resulting expansion and compression ratios.

\begin{table}[t]
\centering
\caption{
\textbf{Effective-dimensionality robustness to representation
normalization.}
The expansion--compression structure persists after removing sample-level
norm, feature-wise scale, or covariance effects.
}
\label{tab:ed_normalization_control}
\vspace{2mm}
\small
\begin{tabular}{lcc}
\toprule
Representation
& $\rho_{\mathrm{exp}}$
& $\rho_{\mathrm{com}}$ \\
\midrule
Raw
& 4.00$\times$
& 2.41$\times$ \\

Row normalization
& 3.70$\times$
& 2.35$\times$ \\

Column standardization
& 3.16$\times$
& 1.69$\times$ \\

Whitening
& 4.00$\times$
& 2.41$\times$ \\
\bottomrule
\end{tabular}
\end{table}

All six tested datasets retain expansion and compression under every
normalization control. Row normalization, in particular, leaves the ratios
close to the raw values ($3.70\times/2.35\times$ versus
$4.00\times/2.41\times$), showing that the late ED reduction is not
explained by representation-norm growth alone.

\paragraph{Random-weight control.}

We next test whether expansion followed by compression arises automatically
from repeatedly applying the TabPFN-3 architecture, independent of learned
weights. On representative datasets, we compare the pretrained model
with two controls: one in which only the ICL blocks are randomly
reinitialized while the surrounding embedding and decoder components remain
pretrained, and one in which all model weights are randomly initialized.
For the pretrained model, the four datasets exhibit expansion ratios between
$2.88\times$ and $7.50\times$ and compression ratios between
$2.09\times$ and $3.33\times$. In both randomized conditions, all four
datasets instead yield
\begin{equation}
\rho_{\mathrm{exp}}
=
\rho_{\mathrm{com}}
=
1.00,
\end{equation}
corresponding to essentially flat ED trajectories.
On these four datasets, randomizing either the ICL blocks or all weights
therefore removes the expansion--compression trajectory. This control is
limited to the tested settings and is not a claim about arbitrary randomly
initialized Transformers.

\paragraph{Robustness of the representative phase boundaries.}

L9 and L17 are representative analytical boundaries derived from the
aggregate ED trajectory, not estimated change points. For the sample-level
trajectory analysis, we also replace the default $(L9,L13)$ pair with
$(L8,L12)$, $(L10,L14)$, $(L9,L14)$, and $(L8,L13)$; the qualitative
trajectory structure is preserved under these nearby choices.

\paragraph{Joint-fit PCA displacement.}

PCA coordinates are identifiable only up to rotations and sign changes.
Fitting PCA independently at each layer could therefore cause a change in
coordinate basis to be mistaken for movement of the underlying
representations. To avoid this artifact, our PCA-displacement analysis uses
a single shared basis. We concatenate query representations from all 24
layers and fit PCA once, retaining
\begin{equation}
k = \min(10,d,n-1)
\end{equation}
components. Every layer is then projected into this same coordinate system.
If $z_t^{(\ell)}$ denotes the resulting projection of sample $t$ at layer
$\ell$, adjacent-layer displacement is computed as
\begin{equation}
d_{\mathrm{PCA}}
\left(
H^{(\ell)},H^{(\ell+1)}
\right)
=
\mathbb{E}_{t\in\mathcal{T}}
\left[
\left\|
z_t^{(\ell+1)}
-
z_t^{(\ell)}
\right\|_2
\right].
\end{equation}
Displacement is computed within each dataset before equal-weight
cross-dataset aggregation. Using one shared PCA basis prevents independent
layerwise basis rotations from contributing to the measured displacement.

\section{Candidate Competition, Early Readability, and Late Resolution}
\label{app:candidate_resolution}

This appendix provides the full diagnostic analyses supporting
Section~\ref{sec:candidate_resolution}. We first define the multiclass task
pool and the prototype-based candidate measures used throughout the
analysis, then report supplementary trajectories of candidate competition
across depth, examine when class-relevant information becomes linearly
readable from the attention-aggregated value stream, and finally quantify
the coupling between late dimensional compression and candidate resolution.

\subsection{Task Pool and Candidate Diagnostics}
\label{app:candidate_tasks}

The analysis covers all 105 multiclass classification tasks used in the
main-text candidate study, including 25 tasks from OpenML-CC18 and 80 tasks
from TALENT. All tasks are evaluated with the same frozen TabPFN-3 model
and the same layerwise representation-extraction protocol.

All diagnostic quantities are first computed at the query level, then
averaged within each dataset, and finally aggregated across datasets with
equal dataset weight, so that datasets with more query examples do not
dominate the pooled statistics.

For the difficulty-stratified analysis in
Figure~\ref{fig:candidate_resolution}(b), tasks are grouped according to
final full-model classification accuracy. Tasks with final accuracy at least
$90\%$ are labeled \emph{Low difficulty}, tasks with final accuracy between
$70\%$ and $90\%$ are labeled \emph{Medium difficulty}, and tasks below
$70\%$ are labeled \emph{High difficulty}.

For each layer $\ell$ and support class $c$, we construct a layer-specific
prototype by averaging the support representations belonging to that class:
\begin{equation}
\mu_c^{(\ell)}
=
\frac{1}{|\mathcal{S}_c|}
\sum_{n \in \mathcal{S}_c}
h_{n,\mathrm{s}}^{(\ell)},
\end{equation}
where $\mathcal{S}_c$ denotes the support samples of class $c$.

For each query $i$, we compute the cosine similarity between its layer-$\ell$
representation and every class prototype:
\begin{equation}
s_{i,c}^{(\ell)}
=
\cos\!\left(h_i^{(\ell)}, \mu_c^{(\ell)}\right).
\end{equation}

Our primary measure of candidate separation is the margin between the two
most similar class prototypes:
\begin{equation}
\Delta_i^{(\ell)}
=
s_{i,(1)}^{(\ell)} - s_{i,(2)}^{(\ell)},
\end{equation}
where
$s_{i,(1)}^{(\ell)} \ge s_{i,(2)}^{(\ell)}$
are the largest and second-largest prototype similarities. A small margin
indicates that the two leading class-aligned candidates remain closely
matched, whereas a large margin indicates more decisive separation.

We additionally record the rank of the correct-class prototype and the
\emph{correct-class top-rank rate}, defined as the fraction of queries for
which the correct prototype occupies rank one.

To characterize the broader similarity structure across all class-aligned
candidates, we also retain two supplementary diagnostics. First, we convert
the prototype similarities into a normalized candidate distribution
\begin{equation}
\pi_{i,c}^{(\ell)}
=
\frac{\exp(s_{i,c}^{(\ell)})}
{\sum_{c'} \exp(s_{i,c'}^{(\ell)})},
\end{equation}
and define the normalized candidate entropy
\begin{equation}
\widetilde{\mathcal{H}}_i^{(\ell)}
=
-
\frac{
\sum_c \pi_{i,c}^{(\ell)} \log \pi_{i,c}^{(\ell)}
}{
\log C_i
},
\end{equation}
where $C_i$ is the number of classes in the corresponding task.

Second, we define the active candidate fraction
\begin{equation}
A_i^{(\ell)}
=
\frac{1}{C_i}
\sum_c
\mathbf{1}
\!\left[
s_{i,c}^{(\ell)}
>
0.6 \max_{c'} s_{i,c'}^{(\ell)}
\right].
\end{equation}

This quantity measures the fraction of class prototypes that remain
competitive under a fixed relative-similarity threshold. These two
quantities serve as supplementary consistency checks rather than the
primary basis of the candidate-resolution conclusion.

\subsection{Candidate Competition Across Depth}
\label{app:candidate_diagnostics}

Table~\ref{tab:candidate_layerwise_summary} summarizes the candidate
diagnostics at representative layers.

Across tasks, the endpoint change from L0 to L23 is highly consistent with
candidate resolution: normalized candidate entropy decreases, active
candidate fraction decreases, prototype margin increases, mean correct-class
rank improves, and correct-class top-rank rate increases. The
intermediate-depth trajectories, however, are not strictly monotonic.

Prototype margin increases from $0.055$ at L0 to $0.125$ at L9, then
decreases to $0.104$ at L13 before increasing again and reaching $0.247$ at
L23. Similarly, the active candidate fraction decreases from $0.932$ at L0
to $0.486$ at L13, but rises to $0.529$ at L17 before decreasing again.
Intermediate depth is therefore characterized by persistent candidate
competition together with local reorganization rather than by smooth
monotonic separation.

\begin{table}[t]
\centering
\caption{
\textbf{Candidate diagnostics across the 105 multiclass tasks.}
Entries are equal-weight dataset means. The final column counts tasks whose
L23--L0 change matches the expected direction under candidate resolution.
}
\label{tab:candidate_layerwise_summary}
\small
\setlength{\tabcolsep}{4pt}
\begin{tabular}{lrrrrrrc}
\toprule
Metric
& L0 & L9 & L13 & L17 & L20 & L23
& Consistent \\
\midrule
Normalized candidate entropy
& .997 & .994 & .995 & .993 & .992 & .978
& 104/105 \\
Active candidate fraction
& .932 & .599 & .486 & .529 & .465 & .465
& 103/105 \\
Prototype margin
& .055 & .125 & .104 & .129 & .147 & .247
& 104/105 \\
Mean correct-class rank
& 2.082 & 1.753 & 1.803 & 1.727 & 1.689 & 1.645
& 104/105 \\
Correct-class top-rank rate
& .634 & .731 & .710 & .729 & .739 & .770
& 104/105 \\
\bottomrule
\end{tabular}
\end{table}

\begin{figure}[t]
    \centering
    \includegraphics[width=\textwidth]
    {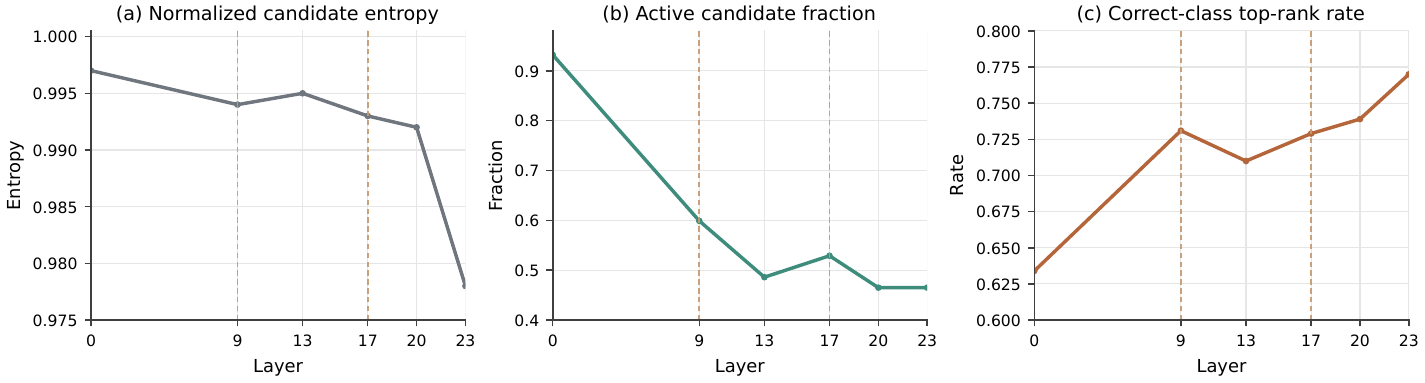}
    \caption{
    \textbf{Supplementary diagnostics of candidate structure across depth.}
    \textbf{(a)} Normalized candidate entropy remains high through
    intermediate depth and decreases most visibly near the output.
    \textbf{(b)} The active candidate fraction contracts overall but shows a
    local rebound in the middle regime.
    \textbf{(c)} Correct-class top-rank rate improves overall, but its
    intermediate trajectory is non-monotonic.
    Dashed vertical lines mark the representative analytical boundaries at
    L9 and L17.
    }
    \label{fig:app_candidate_diagnostics}
\end{figure}

Figure~\ref{fig:app_candidate_diagnostics} visualizes the three
supplementary diagnostics not shown directly in the main text. Candidate
entropy remains close to its maximum through much of the network and drops
most clearly near the output. The active candidate fraction contracts
substantially at early depth but shows a local rebound in the middle
layers. Correct-class top-rank rate also exhibits a mild mid-depth reversal
before improving again toward L23. These observations reinforce the main
text conclusion that candidate resolution is strongly depth-dependent but
not smoothly monotonic.

The strongest candidate separation is concentrated near the output.
Between L20 and L23, prototype margin increases from $0.147$ to $0.247$,
while normalized candidate entropy decreases from $0.992$ to $0.978$.
This late-depth transition coincides with the late effective-dimensionality
contraction and increased PCA displacement reported in
Section~\ref{sec:regime_dynamics}. Candidate geometry therefore remains
only partially resolved through intermediate depth and becomes substantially
more decisive near the output.

The difficulty-stratified correct-class rank curves in
Figure~\ref{fig:candidate_resolution}(b) provide a complementary view.
Lower-difficulty tasks establish favorable correct-class alignment earlier,
whereas higher-difficulty tasks retain worse correct-class ranks deeper into
the network. Candidate ambiguity therefore varies with both depth and task
difficulty.

\subsection{Early Readability of the Attention-Aggregated Value Stream}
\label{app:attn_readability}

We next examine whether persistent candidate ambiguity reflects the absence
of class-relevant information at shallow depth. To do so, we probe the
attention-aggregated value representation at every layer.

For each layer, we train a linear probe on the value representation
produced by learned attention-weighted aggregation, denoted \emph{attn@V}.
As a matched baseline, we apply the same readout to value representations
aggregated with uniform attention weights, denoted \emph{uniform@V}. This
analysis asks when class-relevant information becomes linearly readable,
rather than whether the probe reproduces the model's native decision rule.

\begin{figure}[t]
    \centering
    \includegraphics[width=\textwidth]
    {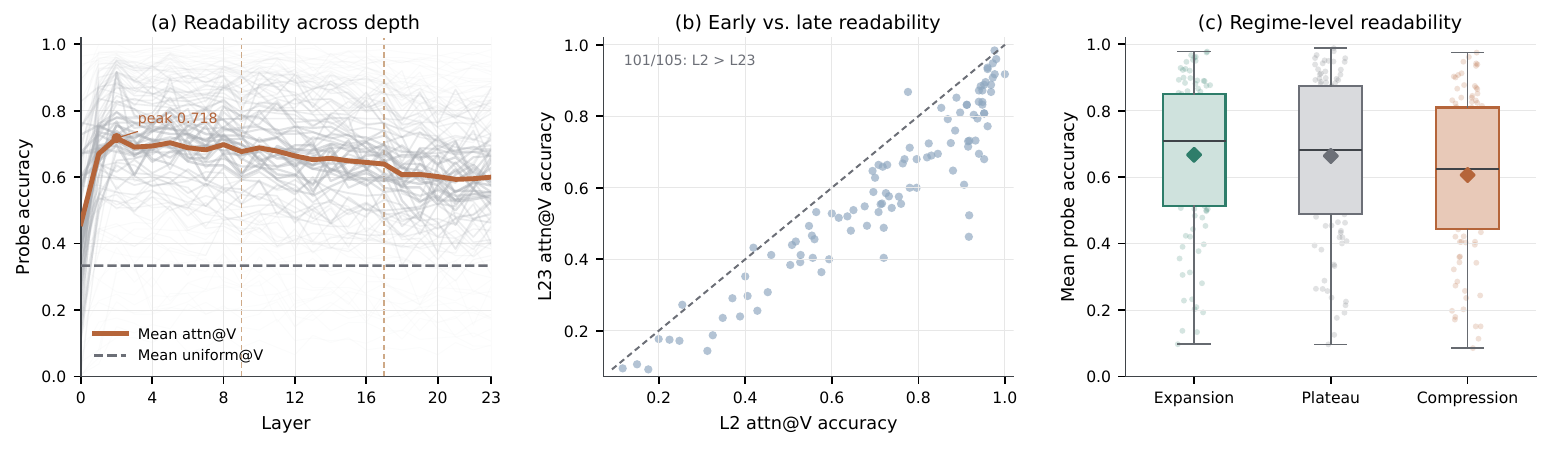}
    \caption{
    \textbf{Class information becomes linearly readable before candidate
    resolution is complete.}
    \textbf{(a)} Mean attn@V probe accuracy across depth, with 95\% bootstrap
    confidence intervals over datasets; uniform@V serves as a matched
    aggregation baseline. The mean attn@V trajectory peaks at L2.
    \textbf{(b)} Dataset-level comparison of attn@V readability at L2 and
    L23. Most datasets lie below the diagonal (101/105), indicating higher
    readability at L2 than at L23.
    \textbf{(c)} Distribution of mean probe accuracy across datasets within
    the Expansion, Plateau, and Compression regimes; diamonds indicate
    equal-weight dataset means.
    }
    \label{fig:app_attn_readability}
\end{figure}

Figure~\ref{fig:app_attn_readability}(a) shows that linear readability
emerges at shallow depth. Mean attn@V accuracy increases from $0.460$ at L0
to $0.718$ at L2, then gradually decreases to $0.600$ at L23. The mean
uniform@V baseline remains near $0.334$ throughout. Class information is
therefore highly linearly readable in the value stream well before candidate
geometry reaches its strongest separation.

This early-readability pattern is also prevalent at the dataset level.
Figure~\ref{fig:app_attn_readability}(b) compares attn@V accuracy at L2 and
L23 for each of the 105 tasks. In 101 of 105 datasets, attn@V accuracy is
higher at L2 than at L23. Figure~\ref{fig:app_attn_readability}(c) shows the
same pattern after regime-wise aggregation. Mean readability is $0.667$ in
the expansion regime, $0.664$ in the plateau regime, and $0.607$ in the
compression regime. Accordingly, 101 of 105 tasks satisfy
Plateau $>$ Compression, whereas no systematic ordering exists between
Expansion and Plateau.

These results separate \emph{information availability} from
\emph{candidate resolution}: class-relevant information becomes linearly
readable early, while candidate geometry continues to reorganize and reaches
its strongest separation substantially later.

\begin{table}[t]
\centering
\caption{
\textbf{Summary statistics for attn@V readability across the 105
multiclass tasks.}
Unless otherwise stated, values are equal-weight dataset means.
}
\label{tab:attn_readability_summary}
\small
\setlength{\tabcolsep}{5pt}
\begin{tabular}{lc}
\toprule
Statistic & Value \\
\midrule
L0 attn@V accuracy & .460 \\
L2 attn@V accuracy & .718 \\
L23 attn@V accuracy & .600 \\
Mean uniform@V accuracy & .334 \\
Expansion mean & .667 \\
Plateau mean & .664 \\
Compression mean & .607 \\
Tasks with L2 $>$ L23 & 101/105 \\
Tasks with Plateau $>$ Compression & 101/105 \\
Tasks with Expansion $>$ Compression & 98/105 \\
\bottomrule
\end{tabular}
\end{table}

\subsection{Coupling Late Compression to Candidate Resolution}
\label{app:compression_resolution_coupling}

We next quantify whether the late geometric contraction identified in
Section~\ref{sec:regime_dynamics} systematically corresponds, across
datasets, to the late increase in candidate separation.

For datasets with both effective-dimensionality and prototype-margin
measurements available over L20--L23, we define relative late compression as
\begin{equation}
C_d
=
\frac{
\mathrm{ED}_d^{(20)} - \mathrm{ED}_d^{(23)}
}{
\mathrm{ED}_d^{(20)}
},
\end{equation}
and late candidate resolution as
\begin{equation}
R_d
=
\Delta_d^{(23)} - \Delta_d^{(20)}.
\end{equation}

\begin{figure}[t]
    \centering
    \includegraphics[width=0.9\textwidth]
    {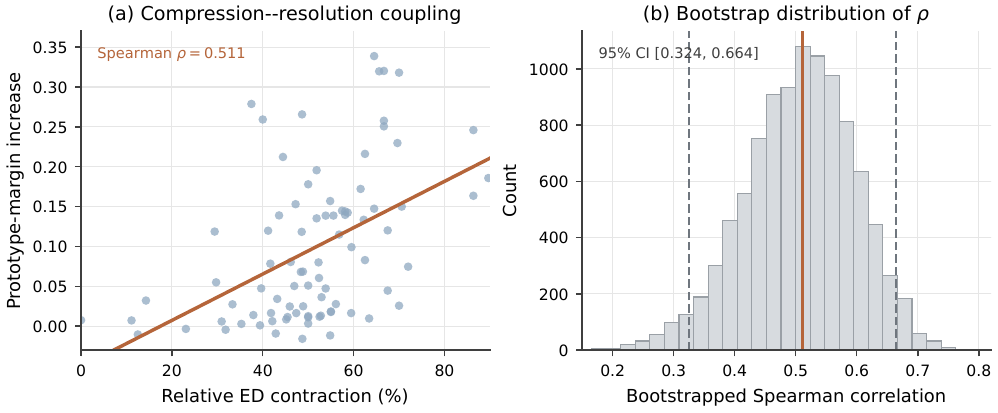}
    \caption{
    \textbf{Robustness of the compression--resolution relationship.}
    \textbf{(a)} Relative effective-dimensionality contraction from L20 to
    L23 versus the corresponding increase in prototype margin across matched
    datasets.
    \textbf{(b)} Bootstrap distribution of the Spearman correlation. The
    solid line marks the observed correlation and dashed lines denote the
    95\% bootstrap confidence interval.
    }
    \label{fig:app_compression_resolution}
\end{figure}

Across all 105 datasets, stronger late compression is associated with larger
late increases in prototype margin, with Spearman
$\rho = 0.511$, $p = 5.85 \times 10^{-7}$, and bootstrap 95\% CI
$[0.324, 0.664]$, as shown in
Figure~\ref{fig:app_compression_resolution}(a). Late compression and
candidate separation therefore systematically covary across datasets.

The relationship remains stable under additional checks. In a
leave-one-dataset-out recomputation, the Spearman correlation remains in the
range $[0.495, 0.546]$. When the late margin increase is normalized by the
L20 prototype margin, the association remains positive, with
$\rho = 0.440$ and $p = 2.56 \times 10^{-5}$. The bootstrap distribution in
Figure~\ref{fig:app_compression_resolution}(b) is likewise concentrated
above zero.

These analyses establish a systematic association between late compression
and late candidate resolution across datasets; the analysis is
correlational and does not identify a causal direction.

\begin{table}[t]
\centering
\caption{
\textbf{Compression--resolution coupling and robustness checks.}
}
\label{tab:compression_resolution_robustness}
\small
\setlength{\tabcolsep}{5pt}
\begin{tabular}{lc}
\toprule
Statistic & Value \\
\midrule
Matched datasets & 105 \\
Analysis interval & L20--L23 \\
Spearman $\rho(C_d, R_d)$ & .511 \\
$p$-value & $5.85 \times 10^{-7}$ \\
Bootstrap 95\% CI & $[.324, .664]$ \\
Leave-one-dataset-out $\rho$ range & $[.495, .546]$ \\
Relative-margin normalization $\rho$ & .440 \\
Relative-margin normalization $p$-value & $2.56 \times 10^{-5}$ \\
\bottomrule
\end{tabular}
\end{table}

\section{Supplementary Analyses of Sample-Level Predictive Commitment}
\label{app:sample_commitment}

This appendix provides the definitions, aggregation details, and robustness
analyses underlying the sample-level results in
Sections~\ref{sec:sample_commitment}--\ref{sec:commitment_difficulty}.
The prediction-trajectory analyses cover 272 datasets. All trajectory
quantities are computed from the complete sequence of 24 native-decoder
predictions rather than inferred from a small set of representative layers.

\subsection{Stabilization Depth and Prediction-Trajectory Assignment}
\label{app:trajectory_patterns}

For sample $i$, let
\begin{equation}
\mathbf{p}_i
=
(\hat y_i^{(0)},\ldots,\hat y_i^{(23)})
\end{equation}
denote the full prediction trajectory. We define stabilization depth as
\begin{equation}
s_i
=
\min
\left\{
\ell:
\hat y_i^{(k)}=\hat y_i^{(23)},
\ \forall k\geq\ell
\right\}.
\end{equation}
Because the final layer always satisfies this condition,
$s_i\in[0,23]$. Samples whose prediction equals the final class at every
layer have $s_i=0$.

We separately define the total number of adjacent-layer class transitions as
\begin{equation}
n_{\mathrm{chg},i}
=
\sum_{\ell=1}^{23}
\mathbb{I}
\!\left[
\hat y_i^{(\ell)}
\neq
\hat y_i^{(\ell-1)}
\right].
\end{equation}

The five-pattern taxonomy used in the main paper is label-free. With
$p_i^{(\ell)}=\hat y_i^{(\ell)}$, the assignment is
\begin{equation}
\operatorname{Pattern}(i)
=
\begin{cases}
\mathrm{ES},
& p_i^{(9)}=p_i^{(23)}
  \ \land\ n_{\mathrm{chg},i}\leq 1,\\
\mathrm{TS},
& p_i^{(9)}=p_i^{(23)}
  \ \land\ n_{\mathrm{chg},i}>1,\\
\mathrm{IS},
& p_i^{(9)}\neq p_i^{(23)}
  \ \land\ p_i^{(13)}=p_i^{(23)},\\
\mathrm{LS},
& p_i^{(9)}=p_i^{(13)}\neq p_i^{(23)},\\
\mathrm{MS},
& \text{otherwise}.
\end{cases}
\label{eq:app_five_patterns}
\end{equation}

The conditions are mutually exclusive and exhaustive. Final correctness is
not used anywhere in the five-pattern assignment. The trajectories shown in
Figure~\ref{fig:sample_dynamics}(b) are schematic representatives. Pattern
assignment uses $n_{\mathrm{chg},i}$ from all 24 layers together with the
representative predictions at L9, L13, and L23; L0 is shown only for visual
context.

\begin{table}[t]
\centering
\caption{
\textbf{Five prediction-transition patterns.}
The rules are applied to the complete 24-layer trajectory and do not use the
ground-truth label.
}
\label{tab:app_pattern_rules}
\small
\setlength{\tabcolsep}{4pt}
\begin{tabular}{lll}
\toprule
Pattern & Representative form & Assignment criterion \\
\midrule
ES
& $A\!\to\!A\!\to\!A\!\to\!A$
& $p^{(9)}=p^{(23)},\ n_{\rm chg}\leq1$ \\
TS
& $A\!\to\!B\!\to\!A\!\to\!A$
& $p^{(9)}=p^{(23)},\ n_{\rm chg}>1$ \\
IS
& $A\!\to\!B\!\to\!B\!\to\!B$
& $p^{(9)}\neq p^{(23)},\ p^{(13)}=p^{(23)}$ \\
LS
& $A\!\to\!A\!\to\!A\!\to\!B$
& $p^{(9)}=p^{(13)}\neq p^{(23)}$ \\
MS
& $A\!\to\!B\!\to\!A\!\to\!C$
& otherwise \\
\bottomrule
\end{tabular}
\end{table}

\subsection{Aggregation Protocol and Pattern Prevalence}
\label{app:pattern_aggregation}

Figure~\ref{fig:sample_dynamics}(b) reports pooled prevalence across all
105,801 test samples. Dataset-level analyses instead assign equal weight to
each of the 272 datasets. Table~\ref{tab:app_pattern_prevalence} shows that
the two aggregation schemes yield similar central prevalences, indicating
that the dominance of ES and the relative rarity of MS are also present
under dataset-level aggregation.

\begin{table}[t]
\centering
\caption{
\textbf{Pattern prevalence under pooled and dataset-macro aggregation.}
Macro values are mean $\pm$ standard deviation across 272 datasets.
}
\label{tab:app_pattern_prevalence}
\small
\begin{tabular}{lcc}
\toprule
Pattern & Pooled samples & Dataset macro-average \\
\midrule
ES & 72.2\% & 71.6\% $\pm$ 22.3\% \\
TS & 15.8\% & 15.9\% $\pm$ 10.9\% \\
IS & 5.3\% & 5.3\% $\pm$ 5.2\% \\
LS & 5.8\% & 6.1\% $\pm$ 8.0\% \\
MS & 1.0\% & 1.1\% $\pm$ 3.5\% \\
\bottomrule
\end{tabular}
\end{table}

For the difficulty visualization in Figure~\ref{fig:sample_dynamics}(c), we
first compute chance-normalized difficulty continuously for every dataset and
then divide datasets into equal-sized tertiles. Pattern fractions are
averaged within each tertile with equal dataset weight.
Table~\ref{tab:app_difficulty_composition} gives the exact values underlying
the stacked bars.

\begin{table}[t]
\centering
\caption{
\textbf{Pattern composition across chance-normalized difficulty tertiles.}
Tertiles are used only for visualization.
}
\label{tab:app_difficulty_composition}
\small
\setlength{\tabcolsep}{4pt}
\begin{tabular}{lccccc}
\toprule
Difficulty & ES & TS & IS & LS & MS \\
\midrule
Easy   & 86.3\% & 9.5\%  & 2.1\% & 2.0\%  & 0.2\% \\
Medium & 74.5\% & 15.0\% & 4.0\% & 5.4\%  & 1.1\% \\
Hard   & 54.2\% & 23.2\% & 9.8\% & 10.9\% & 1.9\% \\
\bottomrule
\end{tabular}
\end{table}

\subsection{Robustness to Representative-Layer Choice}
\label{app:pattern_boundary_robustness}

Stabilization depth is computed from the full trajectory and therefore does
not depend on L9 or L13. The discrete taxonomy uses these depths only to
localize the timing of a transition. To test sensitivity to this choice, we
recompute the five-pattern assignment after shifting the representative
early/intermediate depths by one layer.

Table~\ref{tab:app_layer_sensitivity} shows that every pattern changes by
less than one percentage point from the default prevalence under all tested
alternatives. ES remains dominant, and the relative prevalence of the
remaining categories is qualitatively stable.

\begin{table}[t]
\centering
\caption{
\textbf{Sensitivity of pooled pattern prevalence to representative layers.}
L23 is fixed as the final layer.
}
\label{tab:app_layer_sensitivity}
\small
\setlength{\tabcolsep}{4pt}
\begin{tabular}{lccccc}
\toprule
Representative depths & ES & TS & IS & LS & MS \\
\midrule
Default    & 72.2\% & 15.8\% & 5.3\% & 5.8\% & 1.0\% \\
$L8,L12$  & 72.1\% & 15.5\% & 4.9\% & 6.5\% & 1.1\% \\
$L10,L14$ & 72.3\% & 16.7\% & 5.0\% & 5.0\% & 1.0\% \\
$L9,L14$  & 72.2\% & 15.8\% & 6.1\% & 4.9\% & 1.1\% \\
$L8,L13$  & 72.1\% & 15.5\% & 5.9\% & 5.5\% & 1.0\% \\
\bottomrule
\end{tabular}
\end{table}

\subsection{Task-Difficulty Definition and Class-Imbalance Robustness}
\label{app:difficulty_robustness}

The main text defines model-relative task difficulty using chance-normalized
final accuracy,
\begin{equation}
D_d^{\rm chance}
=
1-
\frac{a_d-1/C_d}
     {1-1/C_d},
\end{equation}
where $a_d$ is final-layer accuracy and $C_d$ is the number of classes.
This normalization removes the first-order dependence of raw accuracy on
class count. All correlations treat difficulty as continuous; tertiles are
used only for visualization.

Because $1/C_d$ is not an imbalance-aware baseline, we additionally test
three alternative definitions: raw error $1-a_d$, balanced error
$1-\mathrm{BA}_d$, and majority-normalized difficulty
\begin{equation}
D_d^{\rm maj}
=
1-
\frac{a_d-b_d}
     {1-b_d},
\end{equation}
where $b_d$ is the majority-class frequency in the evaluation split.

\begin{figure}[t]
\centering
\includegraphics[width=0.8\columnwidth]
{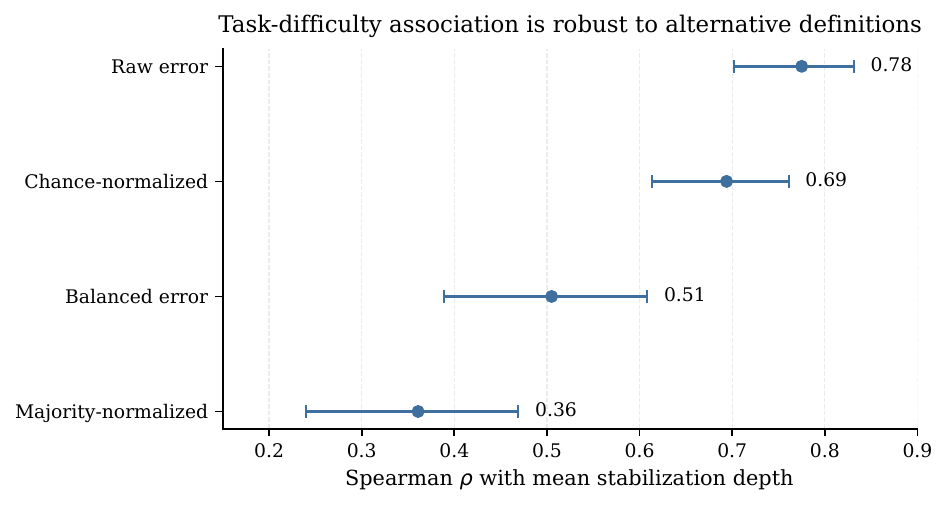}
\caption{
\textbf{Difficulty--commitment association under alternative task
difficulty definitions.}
Points show Spearman correlations with mean stabilization depth; error bars
are 95\% dataset-bootstrap intervals.
}
\label{fig:app_difficulty_robustness}
\end{figure}

Table~\ref{tab:app_difficulty_robustness} and
Figure~\ref{fig:app_difficulty_robustness} report the association of each
measure with mean stabilization depth. The association remains positive
under all four definitions. Its magnitude is strongest for raw and
class-count-normalized difficulty and remains statistically resolved under
both imbalance-aware alternatives.

\begin{table}[t]
\centering
\caption{
\textbf{Robustness of the difficulty--commitment association.}
Bootstrap intervals resample datasets (5,000 replicates). Positive $\rho$
means harder datasets stabilize later on average.
}
\label{tab:app_difficulty_robustness}
\small
\setlength{\tabcolsep}{4pt}
\begin{tabular}{lcccc}
\toprule
Difficulty definition & $n$ & Spearman $\rho$ & 95\% bootstrap CI & $p$ \\
\midrule
Raw error
& 272 & 0.775 & [0.702, 0.831] & $3.8\times10^{-52}$ \\
Chance-normalized
& 272 & 0.694 & [0.614, 0.761] & $8.2\times10^{-38}$ \\
Balanced error
& 272 & 0.505 & [0.389, 0.608] & $7.5\times10^{-18}$ \\
Majority-normalized
& 272 & 0.361 & [0.240, 0.469] & $3.1\times10^{-9}$ \\
\bottomrule
\end{tabular}
\end{table}

\subsection{Representation Dynamics After Early Predictive Commitment}
\label{app:es_post_commitment_geometry}

The main representation analyses characterize geometry over the full set of
query samples. We further examine whether substantial representation
dynamics persist after predicted class identity has already stabilized. To
this end, we repeat the geometric analysis after restricting the query set
to \emph{Early-Stable (ES)} samples.

Under the trajectory definition in
Appendix~\ref{app:trajectory_patterns}, an ES sample satisfies
$p_i^{(9)}=p_i^{(23)}$ and $n_{\mathrm{chg},i}\leq 1$. These conditions
imply that its predicted class cannot change after L9: a departure from the
L9 prediction followed by a return to the final prediction would require at
least two additional class transitions. L9 therefore provides a common
reference point after which all ES predictions have committed to their
final class identity.

For this targeted analysis, we consider the 60 datasets with the highest
ES prevalence, providing well-populated ES subsets for representation-space
estimation. ES prevalence ranges from 88.2\% to 100\% across these datasets,
corresponding to 23,737 ES queries among 25,426 evaluated samples. This
analysis specifically characterizes post-commitment representation dynamics
in datasets dominated by early predictive commitment.

For each dataset, we recompute effective dimensionality using only ES query
representations while retaining the same definition and 90\% variance
threshold as in Appendix~\ref{app:effective_dimensionality}. We additionally
measure adjacent-layer representation displacement using the shared-PCA
procedure defined in Appendix~\ref{app:pca_displacement}.

\begin{figure}[t]
    \centering
    \includegraphics[width=\textwidth]{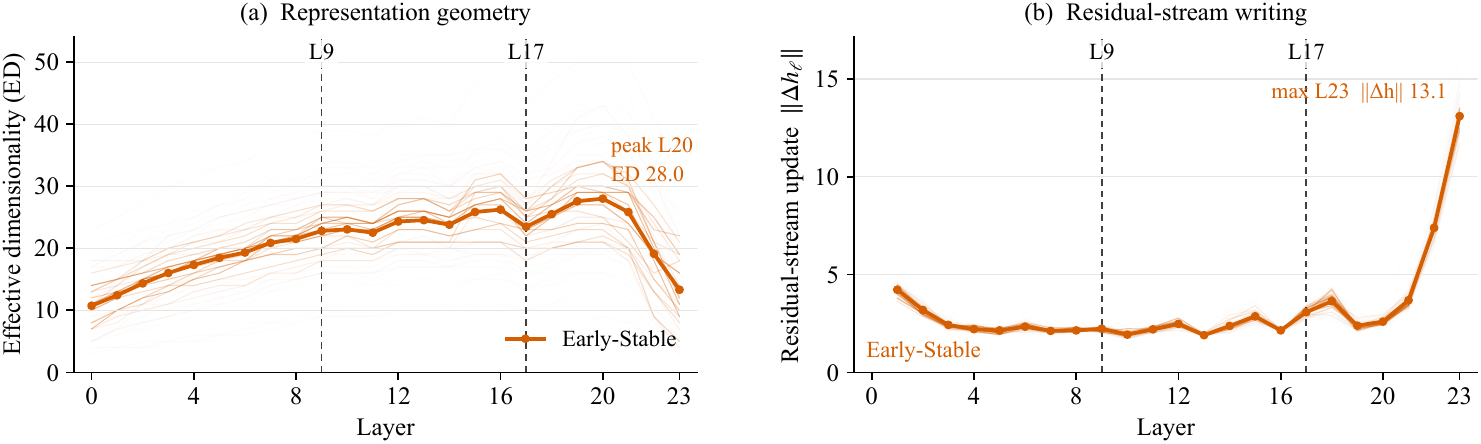}
    \caption{
    \textbf{Representation dynamics continue after predictive commitment.}
    Results are computed using Early-Stable samples only; faded curves denote
    individual datasets and bold curves denote the dataset-level mean.
    \textbf{(a)} Effective dimensionality continues to increase after L9,
    rising from 22.8 at L9 to 28.0 at L20 before contracting to 13.3 at L23.
    \textbf{(b)} Representation movement remains substantial after L9 and is
    concentrated toward later depth.
    }
    \label{fig:app_es_post_commitment_geometry}
\end{figure}

Figure~\ref{fig:app_es_post_commitment_geometry}(a) shows that restricting
the analysis to ES samples preserves the characteristic
Expansion--Compression trajectory. Across the 60 datasets, mean ES-only
effective dimensionality is 22.8 at L9, increases to 28.0 at L20, and then
contracts to 13.3 at L23. In 58 of the 60 datasets, effective dimensionality
rises above its L9 value at a later layer, and all 60 datasets exhibit a
post-L9 dimensional maximum above their final-layer effective
dimensionality.

The displacement analysis provides a complementary view
(Figure~\ref{fig:app_es_post_commitment_geometry}(b)). Despite stable class
identity, ES representations continue to move substantially after L9.
Averaged at the dataset level, 69.2\% of total adjacent-layer displacement
occurs after L9, while the final six transitions alone account for 43.8\%.
Mean transition-wise displacement is also larger in the late regime than in
the intermediate regime (5.47 versus 2.37), with this ordering observed in
all 60 analyzed datasets.

\begin{table}[t]
\centering
\caption{
\textbf{Summary of ES-only post-commitment representation dynamics.}
All statistics are computed after restricting the query representations to
Early-Stable samples. Displacement fractions are first computed within each
dataset and then macro-averaged across datasets.
}
\label{tab:app_es_post_commitment_summary}
\small
\setlength{\tabcolsep}{5pt}
\begin{tabular}{lc}
\toprule
Statistic & ES-only result \\
\midrule
Analyzed datasets & 60 \\
ES queries & 23,737 \\
ES prevalence range & 88.2--100\% \\
Mean ED at L9 & 22.8 \\
Mean ED at L20 & 28.0 \\
Mean ED at L23 & 13.3 \\
Datasets with ED increase after L9 & 58/60 \\
Datasets with post-L9 contraction & 60/60 \\
Displacement occurring after L9 & 69.2\% \\
Displacement in final six transitions & 43.8\% \\
\bottomrule
\end{tabular}
\end{table}

These results separate predictive commitment from representational stasis.
Even after predicted class identity has stabilized, ES representations
continue to undergo substantial depth-dependent transformation, including
further expansion, late reorganization, and final compression. Predictive
commitment can therefore precede substantial later representational change.

\subsection{Local Class Separation and Dataset-Aware Statistics}
\label{app:local_separation_details}

The local-topology analysis is performed in the preprocessed input feature
space, not in a hidden representation. Numerical features follow the common
training-fitted standardization used by the data pipeline, and categorical
features use the same encoded input representation used by the topology
analysis.

For each analyzed sample $x_i$, the diagnostic compares the nearest
same-class and different-class distances,
\begin{equation}
r_i
=
\frac{
d(x_i,\mathrm{NN}_{\rm diff})
}{
d(x_i,\mathrm{NN}_{\rm same})
}.
\end{equation}
Larger $r_i$ indicates clearer local class separation. Class labels are used
only to identify same- versus different-class neighbors for this
retrospective diagnostic; they are not used by the five-pattern trajectory
classifier.

\begin{figure}[!ht]
\centering
\includegraphics[width=0.8\columnwidth]
{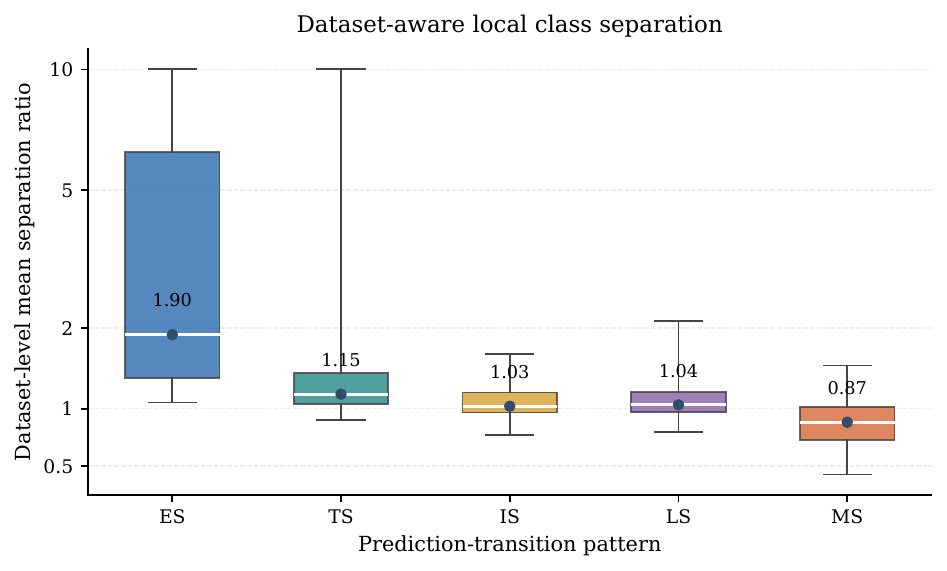}
\caption{
\textbf{Dataset-aware local class separation.}
Each box summarizes within-dataset mean separation ratios for one trajectory
pattern. The vertical axis is log-compressed and values above 10 are clipped
for display; all reported statistics use the uncapped ratios. Tick labels
remain on the original ratio scale. Numbers below the axis indicate datasets
containing a finite estimate for that pattern.
}
\label{fig:app_local_separation_dataset}
\end{figure}

The pooled medians shown in the main text are descriptive. Because samples
are nested within datasets, we additionally perform dataset-aware inference.
For each dataset and trajectory pattern, we first average the available
sample-level separation ratios.
Figure~\ref{fig:app_local_separation_dataset} shows the resulting
distribution of dataset-level pattern means. ES has the largest median
dataset-level separation, whereas MS has the smallest.

\begin{table}[!ht]
\centering
\caption{
\textbf{Local separation at sample and dataset levels.}
Sample entries report pooled median [IQR]. Dataset entries report the median
[IQR] of within-dataset pattern means.
}
\label{tab:app_local_separation_summary}
\small
\setlength{\tabcolsep}{3.5pt}
\begin{tabular}{lrrr}
\toprule
Pattern & Pooled median [IQR] & Datasets & Dataset median [IQR] \\
\midrule
ES & 1.52 [1.11, 2.50] & 270 & 1.90 [1.34, 6.26] \\
TS & 1.03 [0.85, 1.34] & 263 & 1.15 [1.05, 1.39] \\
IS & 0.97 [0.76, 1.16] & 251 & 1.03 [0.96, 1.17] \\
LS & 0.96 [0.72, 1.16] & 251 & 1.04 [0.97, 1.17] \\
MS & 0.81 [0.61, 1.01] & 93  & 0.87 [0.71, 1.02] \\
\bottomrule
\end{tabular}
\end{table}

We further compare ES with each non-ES pattern only within datasets for
which both estimates are available.
Table~\ref{tab:app_local_separation_paired} reports one-sided paired
Wilcoxon tests with Holm correction across the four comparisons. ES has
higher local separation in approximately 95\% of shared datasets for every
comparison, including MS. This dataset-aware analysis reproduces the pooled
ordering without treating individual samples from the same dataset as
independent experimental units.

\begin{table}[!ht]
\centering
\caption{
\textbf{Dataset-aware paired comparisons of local separation.}
$\Delta$ is the median within-dataset difference in mean separation ratio;
$p$ values use a one-sided paired Wilcoxon test with Holm correction.
}
\label{tab:app_local_separation_paired}
\small
\setlength{\tabcolsep}{4pt}
\begin{tabular}{lcccc}
\toprule
Comparison & Shared datasets & Median $\Delta$ & ES higher & Holm $p$ \\
\midrule
ES $>$ TS & 263 & 0.57 & 94.7\% & $<10^{-11}$ \\
ES $>$ IS & 251 & 0.78 & 96.5\% & $<10^{-11}$ \\
ES $>$ LS & 251 & 0.72 & 94.8\% & $<10^{-11}$ \\
ES $>$ MS & 93  & 0.86 & 94.5\% & $<10^{-11}$ \\
\bottomrule
\end{tabular}
\end{table}

\section{Supplementary Analyses of Late-Layer Functional Value}
\label{app:functional_depth}

This appendix provides additional analyses supporting
Section~\ref{sec:functional_depth}. We first examine the robustness of the
relation between predictive commitment and truncation sensitivity across
alternative truncation depths and evaluation metrics. We then test the
stability of the retention--rescue analysis to the choice of shallow
reference layer, followed by geometric, compute-matched, predictive, and
layer-order controls for the functional role of later computation.

\subsection{Robustness of Late-Layer Utility}
\label{app:late_utility}

The main text examines whether tasks with later predictive commitment are
more sensitive to removing subsequent computation. For dataset $d$, the
accuracy loss induced by truncation at layer $\ell$ is defined as
\begin{equation}
\Delta \mathrm{Acc}_d^{(\ell)}
=
\mathrm{Acc}_d^{(23)}
-
\mathrm{Acc}_d^{(\ell)}.
\end{equation}

We repeat the analysis at L13, L17, and L20 and additionally replace
accuracy with balanced accuracy to reduce sensitivity to class imbalance.

Table~\ref{tab:app_truncation_robustness} shows that mean stabilization
depth is positively associated with truncation loss under all six variants.
The association is strongest for earlier truncation points, where a larger
portion of the remaining computation is removed, and remains positive at
L20.

\begin{table}[!ht]
\centering
\caption{
\textbf{Robustness of the commitment--truncation association.}
Spearman correlations relate dataset-level mean stabilization depth to
performance loss under truncation. Confidence intervals are obtained by
dataset bootstrap.
}
\label{tab:app_truncation_robustness}
\small
\setlength{\tabcolsep}{4pt}
\begin{tabular}{lccc}
\toprule
Truncation metric & $\rho$ & 95\% CI & $p$ \\
\midrule
Accuracy, L13
    & .551 & [.443, .643] & $1.55\times10^{-21}$ \\
Balanced accuracy, L13
    & .468 & [.365, .561] & $3.25\times10^{-15}$ \\
Accuracy, L17
    & .481 & [.366, .583] & $3.95\times10^{-16}$ \\
Balanced accuracy, L17
    & .408 & [.300, .504] & $1.32\times10^{-11}$ \\
Accuracy, L20
    & .344 & [.220, .456] & $1.80\times10^{-8}$ \\
Balanced accuracy, L20
    & .308 & [.191, .419] & $5.72\times10^{-7}$ \\
\bottomrule
\end{tabular}
\end{table}

Later computation also improves probabilistic prediction quality.
Comparing L13 with L23, NLL decreases in 58 of 60 datasets (96.7\%), while
ECE decreases in 57 of 60 datasets (95.0\%). Top-1 accuracy increases in 54
datasets, remains unchanged in three, and decreases in three. Improvements
in NLL and ECE are therefore more consistent across tasks than changes in
the discrete class prediction.

We additionally evaluate each depth using a layer-specific matched readout.
Table~\ref{tab:app_matched_readout} reports the resulting trajectory
alongside the native decoder.

\begin{table}[!ht]
\centering
\caption{
\textbf{Native and layer-specific matched readouts across depth.}
A separate matched readout is trained at each layer to evaluate predictive
information available in the corresponding representation.
}
\label{tab:app_matched_readout}
\small
\setlength{\tabcolsep}{3.8pt}
\begin{tabular}{ccccc}
\toprule
Depth &
Native Acc. &
Native NLL &
Matched Acc. &
Matched NLL \\
\midrule
L2  & .804 & .490 & .822 & .459 \\
L5  & .831 & .434 & .832 & .464 \\
L8  & .845 & .409 & .831 & .460 \\
L9  & .848 & .402 & .842 & .414 \\
L13 & .848 & .408 & .847 & .409 \\
L17 & .859 & .396 & .847 & .421 \\
L19 & .856 & .393 & .854 & .410 \\
L21 & .868 & .348 & .842 & .446 \\
L23 & .874 & .284 & .856 & .466 \\
\bottomrule
\end{tabular}
\end{table}

Matched-readout accuracy increases from .822 at L2 to .856 at L23, a
3.4 percentage-point gain. Native accuracy increases from .804 to .874
over the same interval. Later representations therefore provide greater
classification utility even when each layer is evaluated with its own
matched readout.

The probabilistic trajectories differ from the accuracy trajectories.
Native NLL decreases substantially with depth, whereas matched-readout NLL
remains comparatively stable. The native decoder therefore extracts
increasingly better probabilistic predictions from later representations,
while the matched-readout trajectory shows a smaller but persistent increase
in classification accuracy.

\FloatBarrier

\subsection{Sensitivity to the Shallow Reference Layer}
\label{app:shallow-anchor}

The retention--rescue analysis in the main text uses L9 as the shallow
reference layer. We repeat the analysis with L8 and L10 while holding the
remaining protocol fixed.

\begin{figure}[t]
    \centering
    \includegraphics[
        width=\textwidth,
        trim=20 6.0cm 10 8.0cm,
        clip
    ]{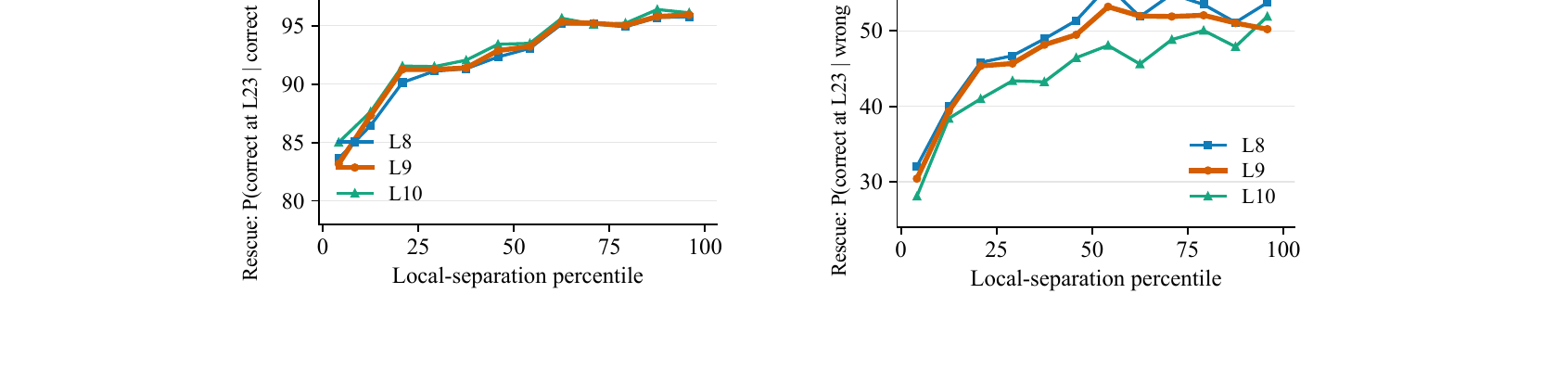}
    \caption{
    \textbf{Retention and rescue across alternative shallow reference
    layers.}
    \textbf{(a)} Retain,
    $P(\text{correct at L23}\mid\text{correct at the reference layer})$,
    and
    \textbf{(b)} Rescue,
    $P(\text{correct at L23}\mid\text{wrong at the reference layer})$,
    plotted against within-dataset local-separation percentile.
    Results are shown for L8, L9, and L10, with L9 used as the main-text
    reference. Curves denote cohort means over 272 datasets.
    }
    \label{fig:shallow-anchor}
\end{figure}
\begin{table}[!ht]
\centering
\caption{
\textbf{Retention and rescue under different shallow reference layers.}
Values are cohort means over 272 datasets. The first two columns summarize
the endpoint changes across local-separation percentiles, and the final two
columns report sample-level Spearman correlations with local separation
($n\!\approx\!9.6\times10^{4}$; all $p<10^{-230}$).
}
\label{tab:shallow-anchor}
\small
\setlength{\tabcolsep}{5pt}
\begin{tabular}{lcccc}
\toprule
Shallow reference &
Retain (\%) &
Rescue (\%) &
$\rho_{\text{retain}}$ &
$\rho_{\text{rescue}}$ \\
\midrule
L8  & 83.6 $\rightarrow$ 95.8 & 32.0 $\rightarrow$ 53.7 & +0.203 & +0.253 \\
L9  & 83.2 $\rightarrow$ 95.9 & 30.4 $\rightarrow$ 50.2 & +0.187 & +0.244 \\
L10 & 85.0 $\rightarrow$ 96.1 & 28.1 $\rightarrow$ 51.9 & +0.194 & +0.240 \\
\bottomrule
\end{tabular}
\end{table}

Across all three reference layers, retention and rescue increase with local
separation, with similar correlation magnitudes and response profiles. The relation between local separation and later predictive change
is therefore stable across nearby shallow reference depths.

\subsection{Shared-Weight and Matched-Compute Controls}
\label{app:shared_weight_controls}

The shared-weight intervention compares the standard sequence of distinct
ICL blocks with repeated application of a fixed transformation. We extract a
single ICL block and repeatedly apply it to the evolving residual stream,
using the same geometric analysis as for the standard 24-layer network.

For a representation trajectory with effective dimension
$\mathrm{ED}^{(\ell)}$, we define the compression ratio as
\begin{equation}
R_{\mathrm{comp}}
=
\frac{
\max_{\ell}\mathrm{ED}^{(\ell)}
}{
\mathrm{ED}^{(\mathrm{final})}
},
\end{equation}
and define FDR gain as the ratio between the final and initial
class-discriminative Fisher ratio.

Table~\ref{tab:app_shared_primary} summarizes the primary comparison.
Repeated application of the shared block produces substantially weaker late
compression and weaker growth of class-discriminative structure.

\begin{table}[!ht]
\centering
\caption{
\textbf{Standard depth-specific computation versus shared-weight
iteration.}
Values are dataset means $\pm$ standard deviations. Wilcoxon tests are
one-sided for the hypothesis that the standard model yields the larger
quantity.
}
\label{tab:app_shared_primary}
\small
\setlength{\tabcolsep}{5pt}
\begin{tabular}{lccc}
\toprule
Metric & Standard & Shared & $p$ \\
\midrule
Compression ratio
    & $2.85\pm1.01$
    & $1.05\pm0.08$
    & $9.5\times10^{-7}$ \\
FDR gain
    & $2.53\pm1.94$
    & $0.84\pm0.26$
    & $2.9\times10^{-6}$ \\
\bottomrule
\end{tabular}
\end{table}

All shared-weight runs have compression ratios below 1.3. The standard
depth-specific sequence therefore exhibits substantially stronger late
compression than repeated application of a fixed block. The FDR comparison
shows a parallel difference in class-discriminative structure, which grows
strongly across the standard sequence but remains much smaller under shared
weights.

We next match the number of computational transformations by repeating the
shared block for 24 steps. Expansion is defined as
\begin{equation}
R_{\mathrm{exp}}
=
\frac{
\max_{\ell}\mathrm{ED}^{(\ell)}
}{
\mathrm{ED}^{(0)}
},
\end{equation}
with compression defined as above.

\begin{table}[!ht]
\centering
\caption{
\textbf{Matched-step shared-weight control.}
Each entry reports expansion/compression ratio. The shared block is
iterated for 24 steps, matching the number of transformations in the
standard network.
}
\label{tab:app_shared24}
\small
\setlength{\tabcolsep}{6pt}
\begin{tabular}{lcc}
\toprule
Dataset & Standard 24-layer & Shared $\times24$ \\
\midrule
vehicle
    & $4.00/2.33$ & $1.57/1.22$ \\
segment
    & $4.67/2.33$ & $1.33/1.14$ \\
satimage
    & $7.50/3.33$ & $2.50/1.00$ \\
diabetes
    & $2.88/2.09$ & $1.62/1.62$ \\
\bottomrule
\end{tabular}
\end{table}

Across all four tasks, the 24-step shared-weight trajectory remains
substantially different from the standard model. Expansion is weaker in
every case, and compression remains below the corresponding standard value.
Distinct depth-specific transformations therefore produce a different
representation trajectory from repeated application of the same operation,
even at matched computational depth.

\subsection{Predictive Robustness Under Data Perturbations}
\label{app:robustness_matrix}

We further compare the standard and shared-weight models under perturbed
evaluation conditions. For each dataset and condition, we report
\begin{equation}
\Delta \mathrm{Acc}
=
\mathrm{Acc}_{\mathrm{standard}}
-
\mathrm{Acc}_{\mathrm{shared}},
\end{equation}
so that positive values favor the standard model.

Table~\ref{tab:app_robustness_matrix} reports results on ten representative
datasets. The clean condition is included alongside 20\% label noise,
feature noise, and reduced training-sample availability.

\FloatBarrier

\begin{table*}[!ht]
\centering
\caption{
\textbf{Predictive advantage of the standard model under data
perturbations.}
Entries are
$\Delta\mathrm{Acc}
=
\mathrm{Acc}_{\mathrm{standard}}
-
\mathrm{Acc}_{\mathrm{shared}}$.
Positive values indicate higher accuracy for the standard model.
}
\label{tab:app_robustness_matrix}
\small
\setlength{\tabcolsep}{7pt}
\begin{tabular}{lrrrr}
\toprule
Dataset &
Clean &
20\% label noise &
Feature noise &
Half samples \\
\midrule
adult
    & +.005 & +.030 &  .000 & +.050 \\
cmc
    & +.030 & +.065 & +.035 & +.090 \\
credit-g
    & +.030 & +.015 & +.040 & +.085 \\
diabetes
    & -.019 & +.045 & +.013 &  .000 \\
balance-scale
    & +.232 & +.288 & +.056 & +.256 \\
vehicle
    & +.312 & +.194 & +.153 & +.171 \\
car
    & +.205 & +.180 & +.070 & +.235 \\
bank-marketing
    & +.015 & +.005 & +.010 & -.015 \\
credit-approval
    & +.007 & +.072 & +.043 & +.029 \\
electricity
    & +.060 & +.010 & +.055 & +.010 \\
\midrule
Mean
    & +.088 & +.090 & +.048 & +.091 \\
Standard $>$ Shared
    & 9/10 & 10/10 & 9/10 & 8/10 \\
\bottomrule
\end{tabular}
\end{table*}

The standard model has a positive mean accuracy advantage in all four
conditions. Under 20\% label noise, the advantage occurs on all ten
datasets. Under feature noise, it occurs on nine datasets with one tie; with
half the training samples, it occurs on eight datasets with one tie.

The shared-weight and standard models therefore remain predictively distinct
under clean and perturbed evaluation conditions, with the largest
cross-dataset consistency observed under label noise.

\FloatBarrier

\subsection{Layer-Order Perturbation Statistics}
\label{app:layer_order_stats}

Section~\ref{sec:layer_order} changes the execution order of the learned
layers while preserving the blocks themselves and the number of
computational steps. We consider both a shuffled ordering and complete
reversal of the learned sequence.

Table~\ref{tab:app_layer_order_stats} summarizes the dataset-level effects
on effective-dimensional compression and predictive accuracy. Wilcoxon
tests are paired and one-sided for the hypothesis that the learned ordering
produces the larger value.

\begin{table}[!ht]
\centering
\caption{
\textbf{Dataset-level effects of layer-order perturbation.}
``Learned better'' reports the number of datasets on which the learned
ordering yields the larger metric. Ties are omitted from this count.
}
\label{tab:app_layer_order_stats}
\small
\setlength{\tabcolsep}{3.5pt}
\begin{tabular}{lcccc}
\toprule
Metric &
Learned &
Perturbed &
Learned better &
$p$ \\
\midrule
Compression, shuffled
    & 1.935 & 1.053 & 60/60 & $8.65\times10^{-7}$ \\
Compression, reversed
    & 1.935 & 1.012 & 60/60 & $8.64\times10^{-7}$ \\
Accuracy, shuffled
    & .806 & .650 & 58/60 & $1.42\times10^{-6}$ \\
Accuracy, reversed
    & .806 & .682 & 56/60 & $5.53\times10^{-6}$ \\
\bottomrule
\end{tabular}
\end{table}

Effective-dimensional compression is lower after both shuffling and reversal
on all 60 datasets. The learned ordering also yields higher accuracy than
the shuffled ordering on 58 datasets and higher accuracy than the reversed
ordering on 56 datasets.

Changing layer order therefore alters both the representation trajectory and
predictive behavior. Together with the shared-weight controls, these results
show that later computation is organized as an ordered sequence of distinct
learned transformations.

\FloatBarrier

\end{document}